\documentclass[sigconf]{acmart}

\usepackage{booktabs}
\usepackage{multirow}
\usepackage{colortbl}
\usepackage{makecell}
\usepackage{array}
\usepackage{tabularx}
\usepackage{algorithm}
\usepackage{algpseudocode}
\usepackage{graphicx}
\usepackage{placeins}
\usepackage{afterpage}
\usepackage{draftwatermark}
\SetWatermarkText{Accepted in CIKM 2026}
\SetWatermarkScale{0.45}
\SetWatermarkLightness{0.88}
\SetWatermarkAngle{45}

\copyrightyear{2026}
\acmYear{2026}
\setcopyright{cc}
\setcctype{by-nc-nd}
\acmConference[CIKM '26] {Proceedings of the 35th ACM International Conference on Information and Knowledge Management}{November 7--11, 2026}{Rome, Italy}
\acmBooktitle{Proceedings of the 35th ACM International Conference on Information and Knowledge Management (CIKM '26), November 7--11, 2026, Rome, Italy}
\acmISBN{979-8-4007-2539-5/2026/11}
\acmDOI{10.1145/3799682.3840827}

\usepackage{etoolbox}
\makeatletter
\patchcmd{\maketitle}
  {\acmConference@shortname, \acmConference@venue}
  {\acmConference@shortname, \acmConference@date, \acmConference@venue}
  {}{}
\makeatother

\newcommand{\DI}{\mathrm{DI}}
\newcommand{\SPD}{\mathrm{SPD}}
\newcommand{\EOPP}{\mathrm{EOpp}}
\newcommand{\EOD}{\mathrm{EOD}}
\newcommand{\TI}{\mathrm{TI}}
\newcommand{\PP}{\mathrm{PP}}
\newcommand{\CAL}{\mathrm{Cal}}
\newcommand{\VFR}{\mathrm{VFR}}

\newenvironment{findingbox}[1]{%
  \par\noindent\textbf{\small Finding: #1}\par
  \footnotesize\itshape\noindent\ignorespaces%
}{%
  \par
}

\AtBeginDocument{\providecommand{\shownote}[1]{}\renewcommand{\shownote}[1]{}}

\title{VFR-Audit: Verdict-Level Reliability for Fairness Audits in Hospital Length-of-Stay Prediction}

\author{Md Jannatul Rakib Joy}
\affiliation{%
  \department{School of Science, Computing and Emerging Technologies}
  \institution{Swinburne University of Technology}
  \city{Hawthorn}
  \state{Victoria}
  \country{Australia}
}
\email{103799644@student.swin.edu.au}

\author{Viet Vo}
\affiliation{%
  \department{School of Science, Computing and Emerging Technologies}
  \institution{Swinburne University of Technology}
  \city{Hawthorn}
  \state{Victoria}
  \country{Australia}
}
\email{vvo@swin.edu.au}

\author{Caslon Chua}
\affiliation{%
  \department{School of Science, Computing and Emerging Technologies}
  \institution{Swinburne University of Technology}
  \city{Hawthorn}
  \state{Victoria}
  \country{Australia}
}
\email{cchua@swin.edu.au}

\ccsdesc[500]{Applied computing~Health informatics}
\ccsdesc[500]{Computing methodologies~Machine learning}
\ccsdesc[300]{Social and professional topics~Computing / technology policy}
\ccsdesc[500]{Information systems~Data mining}

\begin{document}

%% =====================================================================
%% ABSTRACT
%% =====================================================================
\begin{abstract}
Fairness audits in clinical Artificial Intelligence convert continuous fairness metrics into binary pass-or-fail verdicts against operational thresholds, where hospital governance boards, payers, and regulators act on the resulting verdicts. Such audits are repeated over time and across hospital sites, thus the same verdict can flip between pass and fail across audits. Existing uncertainty methods such as Bayesian posteriors, bootstrap confidence intervals, and permutation tests address verdict instability only at the continuous-metric level. Converting metric-level uncertainty into a verdict-stability claim remains a manual step that scales poorly across the (model, metric, attribute) cells an audit covers. Existing uncertainty methods also leave open whether bias-mitigation steps, such as reweighing or per-group threshold shifts, yield a stable passing verdict at the cost of model discrimination measured as AUROC or AUPRC.

To address this verdict-stability gap, we propose VFR-Audit, a framework built around the Verdict Flip Rate (VFR), a scalar bounded between 0 and 0.5 that measures the probability of verdict reversal under stratified bootstrap resampling. VFR-Audit reports VFR alongside three reliability axes, namely within-cohort resampling stability, audit-size sensitivity, and cross-hospital verdict agreement via Fleiss' kappa. The framework was evaluated on 925,128 administrative discharge records from 441 Texas hospitals, against an intersectional adaptation of Kamiran and Calders' reweighing. On the unintervened baseline, the audit detects verdict reversal in 43.5\% of cells in the cross-model audit grid. After the canonical intervention, 11 of 28 cells continue to flip. Intersectional reweighing fails the four-fifths rule on this cohort regardless of regularisation strength, while per-cell threshold shifting is the main effective component of the canonical intervention. VFR translates continuous metric uncertainty into a probability of binary-verdict reversal, and a derived-baseline comparison attributes the verdict outcome to specific intervention components, so audit reliability and intervention effects can be reported together rather than as separate claims.
\end{abstract}

\keywords{algorithmic fairness, clinical AI, audit reliability, verdict stability, length-of-stay prediction, multi-site validation, responsible deployment}

\maketitle

%% =====================================================================
%% 1. INTRODUCTION
%% =====================================================================
\section{Introduction}\label{sec:intro}

Length-of-stay (LOS) prediction supports discharge planning, bed management, and care coordination~\cite{rajkomar2018scalable, jain2024los, cai2025protoehr, georgiev2025healthcare}. As clinical AI systems move closer to deployment, fairness evaluation has become a central requirement for assessing whether model behaviour is acceptable across protected patient subgroups defined by race, sex, ethnicity, and age~\cite{gichoya2022ai, mccradden2024responsible, hasanzadeh2025bias, wells2025fairai, alderman2025standing, lekadir2025futureai}. Recent surveys document the breadth of healthcare-AI fairness concerns and the persistence of subpopulation disparities even in high-performance models~\cite{hassanpour2024oodfair, liu2025scoping, subbaswamy2024afisp, obra2025cdi}. Many fairness audits summarise model behaviour by comparing a continuous fairness metric against a predefined threshold, producing a binary pass-or-fail verdict that hospital governance boards, payers, and regulators read and act upon~\cite{barocas2023fairness, verma2018fairness, schwartz2024ai}.

This thresholded verdict is data-dependent. In practical deployment, hospitals acquire new patient records over time. Demographic composition, subgroup size, and site-level case mix of the audit cohort change~\cite{sendak2020implementation, beede2020human, futoma2020myth, roberts2024monitoring, fiske2025raceethnicity}. A fairness verdict that passes on one cohort may therefore fail on another even when the trained model is unchanged~\cite{mehta2024continuous}. For example, when Disparate Impact is assessed using the four-fifths rule from the U.S.\ Equal Employment Opportunity Commission~\cite{eeoc1979uniform, watkins2024fourfifths}, small changes in cohort composition around the $0.80$ threshold can change the audit conclusion. Two deployed-system audits illustrate the operational stakes. Obermeyer et al.~\cite{obermeyer2019dissecting} audited a commercial risk-prediction tool deployed across major U.S.\ health systems and found that, at the same predicted risk score, Black patients had substantially worse health than White patients, with the corrective audit shifting the Black share of care-program enrolment from $17.7$\,\% to $46.5$\,\%. Omar et al.~\cite{omar2025sociodemo} stress-tested nine large language models on $1{,}000$ emergency-department cases across $32$ sociodemographic variants ($1.7 \times 10^{6}$ outputs) and found that cases labelled as Black, unhoused, or LGBTQIA+ were more frequently routed to urgent care, invasive interventions, or mental-health evaluations despite identical clinical details. The deployment-relevant verdict therefore depended on whether and how the audit was performed.

Modern healthcare-AI systems, including generative and agentic systems, are updated and redeployed more frequently than traditional static models~\cite{azarfar2025multimodal, ghassemi2024genai, templin2025llmframework, lyu2024genai}, making one-time certification insufficient~\cite{birhane2024foundation, liu2025fairnessdrift, mehta2024continuous, feng2022aiqi, roberts2024monitoring}. Governance work likewise treats fairness as a lifecycle monitoring requirement~\cite{mccradden2024responsible, wells2025fairai}. The EU AI Act~\cite{eu2024aiact}, FDA PCCP guidance~\cite{fda2023samd}, and NIST AI RMF~\cite{nist2023airmf} reinforce repeatable post-deployment evidence, while deployment and generalisability work highlights validation across settings~\cite{beede2020human, futoma2020myth, sendak2020implementation}. Existing uncertainty methods quantify metric uncertainty using Bayesian posteriors~\cite{barrainkua2024uncertainty}, bootstrap confidence intervals~\cite{cherian2024fairaudit}, permutation tests~\cite{diciccio2020evaluating}, and sample-size formulae~\cite{singh2023sample}, while mitigation methods modify training or post-processing~\cite{mackin2025safetynet, yang2023drlbias}. Neither directly reports how often the resulting thresholded pass-or-fail verdict reverses across repeated audits.

Accordingly, our problem is to determine whether a thresholded fairness verdict for a fixed trained model remains reproducible under cohort resampling, audit-size variation, and hospital shift.

To address this gap, we propose the \emph{Verdict Flip Rate} (VFR), a scalar measure of binary-verdict instability under stratified bootstrap resampling. VFR is the smaller of the pass-count and the fail-count across $K$ resamples of the audit cohort, normalised by $K$, and bounded in $[0, 0.5]$. A VFR of zero indicates that all $K$ resamples produce the same verdict. A VFR of $0.5$ indicates that the pass-count equals the fail-count, so the audit grid carries no information about which verdict the next resample would return. VFR separates fairness status from verdict reliability. A model can be stably fair, stably unfair, or unstable near the threshold. The choice of bootstrap count $K$ is fixed by a worst-case binomial standard-error argument with concentration-bound support and transfers across cohorts without re-derivation.

To operationalise VFR for repeated deployment audits, we develop the VFR-Audit framework. Axis 1 evaluates within-cohort resampling stability by repeatedly auditing the same trained model on stratified bootstrap resamples. Axis 2 provides audit-size guidance. The axis estimates the minimum cohort size at which a fairness metric becomes stable, defined as the smallest size at which the coefficient of variation of the metric falls below five per cent across thirty repetitions. Axis 3 evaluates cross-hospital verdict agreement using hospital-grouped validation and Fleiss' $\kappa$. Together, the three axes identify whether unreliable fairness conclusions arise from cohort resampling variation, insufficient audit size, or hospital-level heterogeneity. They feed a deterministic decision rule that classifies each (model, metric, attribute) cell into one of four reliability tiers.\footnote{Throughout this paper, ``clinical AI'' refers broadly to machine-learning systems used for patient-risk prediction, hospital operations, and decision support. The LOS task studied here is primarily a hospital-operational prediction task with clinical-decision-support implications. The framework itself is task-agnostic.}

We validate the framework on the Texas-100X cohort, which contains $9.25 \times 10^{5}$ administrative discharge records from $441$ hospitals~\cite{thcic2006pudf}, audited using twelve standard machine-learning classifiers, seven fairness metrics, and four protected attributes ($336$ audit cells per pass). On the unintervened baseline, $146$ of $336$ ($43.5$\,\%) audited cells in the cross-model grid reverse their verdict under stratified bootstrap resampling. On the canonical post-intervention baseline, $11$ of $28$ cells continue to flip even when the point-estimate audit certifies all four protected attributes as fair. A four-baseline comparison shows that intersectional reweighing alone fails to satisfy the four-fifths rule on this dataset regardless of the regularisation strength applied, while per-cell threshold shifting is the component responsible for the all-four-DI pass.

The main contributions of this paper are as follows.
\begin{itemize}
\item We propose the Verdict Flip Rate (VFR), a scalar measure of binary fairness-verdict instability under stratified bootstrap resampling, with a fixed resample-count selection rule motivated by worst-case binomial uncertainty and concentration bounds.
\item We develop a three-axis reliability framework around VFR covering within-cohort resampling, audit-size guidance, and cross-hospital agreement, combined into a four-tier per-cell reliability classification.
\item We empirically demonstrate the framework on Texas-100X, showing that $146$ of $336$ cells flip under bootstrap on the unintervened baseline, $11$ of $28$ flip after the canonical intervention, and per-cell threshold shifting is the component responsible for the all-four-DI pass.
\end{itemize}

%% =====================================================================
%% 2. RELATED WORK
%% =====================================================================
\section{Related Work}\label{sec:related}

This section places the Verdict Flip Rate in the context of three lines of prior work: length-of-stay prediction, fairness audits in clinical AI, and audit-instrument uncertainty quantification.

\begin{table*}[!t]
\caption{LOS-focused literature comparison across six audit-relevant axes. ``Stab.''\ is verdict stability, ``Site''\ is cross-site validation, ``Size''\ is audit-size guidance. ``n/r''\ denotes not reported.}
\label{tab:los-comparison}
\centering
\scriptsize
\setlength{\tabcolsep}{3pt}
\renewcommand{\arraystretch}{1.05}
\begin{tabular}{@{} l l l p{2.0cm} p{2.4cm} p{2.2cm} c c c @{}}
\toprule
\textbf{Study} & \textbf{Dataset} & \textbf{$N$ (unit)} & \textbf{Task} & \textbf{Model} & \textbf{Performance} & \textbf{Stab.} & \textbf{Site} & \textbf{Size} \\
\midrule
Rajkomar et al.~\cite{rajkomar2018scalable}
  & UCSF/UCM (private) & $2.16 \times 10^{5}$ adm.
  & $\mathrm{LOS} > 7$\,d & DL ensemble (LSTM / TANN / boosted time-based stumps)
  & AUROC $0.85$ to $0.86$ & n/r & multi ($2$) & n/r \\
Rocheteau et al.~\cite{rocheteau2021tpc}
  & eICU-CRD (public) & n/r
  & remaining-LOS regr. & Temporal pointwise conv.
  & MAE $1.55$\,d & n/r & n/r & n/r \\
Wu et al.~\cite{wu2021predicting}
  & eICU $\rightarrow$ MIMIC-III & $1.17 \times 10^{5}$ / $4.29 \times 10^{4}$ pts.
  & prolonged ICU LOS & GBDT (best of $4$)
  & AUROC $0.742$ / $0.747$ & n/r & external & n/r \\
Jain et al.~\cite{jain2024los}
  & SPARCS (public) & $2.30 \times 10^{6}$ records
  & LOS regression & Linear regr.\ / CatBoost regr.
  & $R^{2} \approx 0.82$ (NB) & n/r & n/r & n/r \\
Cai et al.~\cite{cai2025protoehr}
  & MIMIC-III \& MIMIC-IV & $5.15 \times 10^{4}$ pts. / $1.67 \times 10^{5}$ visits
  & LOS multi-class & ProtoEHR (proto.\ learning)
  & $+7.4$ F1 (MIMIC-IV LOS) & n/r & n/r & n/r \\
Li et al.~\cite{li2022improving}
  & GWTG-HF registry & $2.10 \times 10^{5}$ patients
  & $\mathrm{LOS} > 5$\,d (HF) & GB + mitigation
  & AUROC $0.83$ & n/r & n/r & n/r \\
Abakasanga et al.~\cite{abakasanga2025equitable}
  & SAIL LD (UK; restricted) & $9.62 \times 10^{3}$ patients
  & LOS prediction (LD) & RF + threshold opt.
  & AUC $0.759$ (M) / $0.756$ (F) & n/r & n/r & n/r \\
\midrule
\rowcolor{gray!15}
\textbf{This study}
  & \textbf{THCIC PUDF (public)} & $\mathbf{9.25 \times 10^{5}}$ \textbf{records}
  & $\mathbf{\mathrm{LOS} > 3}$\,d & \textbf{GBDT (canon., $12$-model panel)}
  & \textbf{AUROC $\mathbf{0.953}$}
  & \boldmath$\VFR$ & \boldmath$K_\mathrm{hosp}{=}20$ & \textbf{min-}$\mathbf{N^{*}}$ \\
\bottomrule
\end{tabular}
\end{table*}

\subsection{Length-of-Stay Prediction}\label{sec:related-los}

Hospital LOS prediction has been studied as binary classification ($\mathrm{LOS} > 3$, $5$, or $7$ days) and as continuous regression. Reported predictive performance is high. Rajkomar et al.~\cite{rajkomar2018scalable} reached AUROC $= 0.85$ to $0.86$ on $\mathrm{LOS} > 7$\,d with a deep-learning ensemble (LSTM, time-aware attention, and boosted time-based decision stumps) over $2.16 \times 10^{5}$ admissions from two academic medical centres. Wu et al.~\cite{wu2021predicting} trained on $1.17 \times 10^{5}$ eICU-CRD ICU patients (internal $70/30$ split, AUROC $0.742$) and externally validated on $4.29 \times 10^{4}$ MIMIC-III patients (AUROC $0.747$) using a gradient-boosted decision tree for prolonged-ICU-LOS prediction. Jaotombo et al.~\cite{jaotombo2022prolonged} analysed $7.32 \times 10^{4}$ hospitalisations from a French university hospital and reached AUC $= 0.810$ with gradient boosting, outperforming random forest, neural network, and logistic regression baselines (all $p < 0.0001$).

In the regression framing, Mekhaldi et al.~\cite{mekhaldi2021comparative} reached $R^2 = 0.94$ and MAE $= 0.44$\,d with gradient boosting on $1.00 \times 10^{5}$ observations from an open hospital-LOS benchmark. Jain et al.~\cite{jain2024los} reported $R^2 = 0.82$ for newborns using linear regression and $R^2 = 0.43$ for non-newborns using CatBoost regression on $2.30 \times 10^{6}$ SPARCS records. Rocheteau et al.~\cite{rocheteau2021tpc} reached MAE $= 1.55$\,d on remaining-LOS regression with a temporal pointwise convolutional network on the eICU benchmark. Recent deep-learning advances include ProtoEHR~\cite{cai2025protoehr}, which evaluates LOS prediction on MIMIC-III ($5.45 \times 10^{3}$ patients, $1.43 \times 10^{4}$ visits) and MIMIC-IV ($5.15 \times 10^{4}$ patients, $1.67 \times 10^{5}$ visits) and reports a $7.4$\,\% F1 improvement for LOS on MIMIC-IV.

Fairness coverage in the LOS literature is sparse. Li et al.~\cite{li2022improving} audited a $2.10 \times 10^{5}$-patient heart-failure cohort on Disparate Impact, Equal Opportunity, Equalised Odds, and Predictive Parity and found that adding social-determinant features reduced subgroup-AUC disparities by $0.022$ to $0.037$ across race strata. Abakasanga et al.~\cite{abakasanga2025equitable} analysed $9.62 \times 10^{3}$ patients with learning disabilities and reported random-forest AUCs of $0.759$ for males and $0.756$ for females. Chesley et al.~\cite{chesley2023racial} analysed $102{,}362$ adults with sepsis and/or acute respiratory failure across $27$ U.S.\ hospitals and found that, after matching, Black patients had longer LOS than White patients by $1.26$\,d for sepsis and $0.97$\,d for acute respiratory failure. None of the surveyed LOS studies reports verdict stability under cohort resampling, cross-hospital verdict agreement through GroupKFold or Fleiss' $\kappa$, or audit-size guidance for the metrics they evaluate.

\subsection{Fairness Audits in Clinical AI}\label{sec:related-fairness}

Group-fairness criteria including Disparate Impact, Statistical Parity Difference, Equal Opportunity, Equalised Odds, the Theil index, Predictive Parity, and Calibration are formalised in the textbook of Barocas, Hardt and Narayanan~\cite{barocas2023fairness} and surveyed in detail by Verma and Rubin~\cite{verma2018fairness} and Mehrabi et al.~\cite{mehrabi2021survey}. Corbett-Davies et al.~\cite{corbettdavies2023measure} synthesise the field through a measure-versus-mismeasure lens, arguing that no single metric is sufficient for governance, in part because the well-known impossibility result prevents simultaneous satisfaction of more than one criterion when group base-rates differ. These criteria have been applied to clinical prediction across domains spanning intensive-care mortality, ED triage, sepsis, readmission, chest-X-ray classification, and dermatology~\cite{meng2022icufairness, davoudi2024fairness, pfohl2021empirical, wang2024readmission, seyyed2021underdiagnosis, ghassemi2024medicine}. Scoping reviews by Mbakwe et al.~\cite{mbakwe2023fairness} and, more recently, Liu et al.~\cite{liu2025scoping} document accelerating uptake but also persistent metric heterogeneity, the narrow focus on bias-relevant attributes, and the dominance of group fairness centered on model-performance equality; Hasanzadeh et al.~\cite{hasanzadeh2025bias} systematically review bias-recognition and mitigation strategies across the AI model lifecycle, from conception through to deployment and longitudinal surveillance. Intersectional fairness, the requirement that fairness hold jointly across combinations of protected attributes~\cite{mbakwe2023fairness, seyyed2021underdiagnosis}, motivates Baseline 2 in our experiment to operate on the joint (race $\times$ age $\times$ sex) cell structure.

Two purpose-built clinical-AI fairness-audit frameworks have appeared. Cachel and Rundensteiner~\cite{cachel2022fins} developed FINS for group-fairness auditing of subset selections such as top-$k$, multi-winner voting, and clustering. Barnard et al.~\cite{barnard2023macaif} developed MACAIF, a clinician-facing dashboard layered on the MLighter adversarial tool. Neither framework defines a scalar measure of binary-verdict instability, and neither has been validated on a state-wide clinical cohort.

\subsection{Audit-Uncertainty and Verdict Stability}\label{sec:related-uncertainty}

Statistical-uncertainty quantification for fairness metrics has three established branches. The Bayesian branch~\cite{barrainkua2024uncertainty} computes posteriors over fairness-metric values. The frequentist branch~\cite{cherian2024fairaudit} computes bootstrap confidence intervals and simultaneous bounds on subgroup performance disparities. The hypothesis-testing branch~\cite{diciccio2020evaluating} tests null hypotheses through permutation $p$-values. A complementary line of work derives sample-size formulae for fairness audits. Singh et al.~\cite{singh2023sample} provide closed-form expressions via the central limit theorem and the delta method.

A separate line of work directly addresses verdict-level stability. Ganesh et al.~\cite{ganesh2023impact} and Cooper et al.~\cite{cooper2024arbitrariness} show that fairness verdicts can change under random training-seed variation; Black et al.~\cite{black2022model} document predictive multiplicity, the phenomenon that competing models with similar accuracy disagree on individual predictions. The conclusion across this work is consistent. Fairness verdicts are not stable in deployment, and the instability is not solely caused by sampling noise. None of these works, however, defines a scalar instrument for binary-verdict instability under stratified bootstrap that is suitable for governance reporting. The present work fills this gap. Across the closest LOS studies summarised in Table~\ref{tab:los-comparison}, none reports verdict stability under cohort resampling, cross-hospital verdict agreement through GroupKFold or Fleiss' $\kappa$, or audit-size guidance, which defines the empirical gap addressed here.

\FloatBarrier

%% =====================================================================
%% 3. PROPOSED METHOD  (with explicit Baselines subsection inside Method)
%% =====================================================================
\section{Proposed Method}\label{sec:method}

\subsection{Notation}\label{sec:method-notation}

Table~\ref{tab:notation} fixes the notation used throughout the paper. To avoid ambiguity, $N$ denotes the number of \emph{discharge records} in an audit cohort, not the number of unique patients. The same record may correspond to one of multiple admissions for a single patient, and the audit instrument operates on the record-level joint distribution.

\begin{table}[t]
\caption{Notation used in the VFR-Audit framework.}
\label{tab:notation}
\centering
\footnotesize
\setlength{\tabcolsep}{3.5pt}
\renewcommand{\arraystretch}{1.10}
\begin{tabular}{@{} l p{3.6cm} p{2.7cm} @{}}
\toprule
\textbf{Symbol} & \textbf{Meaning} & \textbf{Example / range} \\
\midrule
$\mathcal{D}$ & Audit dataset & Discharge records \\
$N$ & Records in $\mathcal{D}$ (not patients) & $10^{3}$ to $10^{6}$ \\
$f, \mathcal{M}$ & Trained model; model panel & XGBoost, $|\mathcal{M}| = 12$ \\
$a, \mathcal{A}$ & Protected attribute; set & Race, sex, eth., age \\
$m, \mathcal{F}$ & Fairness metric; metric set & DI, SPD, EOPP, EOD, TI, PP, Cal \\
$\tau_m$ & Operational threshold for $m$ & $\DI \geq 0.80$ \\
$v(f, \mathcal{D}, m, a)$ & Binary verdict & $\{\textsc{Pass}, \textsc{Fail}\}$ \\
$\mathcal{C}$ & Audit grid $\mathcal{M} \times \mathcal{F} \times \mathcal{A}$ & $336$ cells (this study) \\
$K$ & Bootstrap-resample count (Axis 1) & $500$ \\
$\VFR(c)$ & Verdict Flip Rate of cell $c$ & $[0, 0.5]$ \\
$\sigma(c)$ & Stability margin (SD from $\tau$) & $\geq 0$ \\
$N^{*}(c)$ & Minimum reliable audit size (Axis 2) & Smallest $N$ with $\mathrm{CV} < 0.05$ \\
$N_\mathrm{field}$ & Field-realistic cohort size & $5 \times 10^{4}$ \\
$K_\mathrm{hosp}$ & Hospital-fold count (Axis 3) & $20$ \\
$\kappa(c)$ & Metric-level Fleiss' $\kappa$ assigned to $c$ & $[-1, 1]$, Landis-Koch bands \\
\bottomrule
\end{tabular}
\end{table}

\subsection{Design Intuition and Deployment Rationale}\label{sec:method-intuition}
\begin{figure}[t]
\centering
\includegraphics[width=0.82\columnwidth]{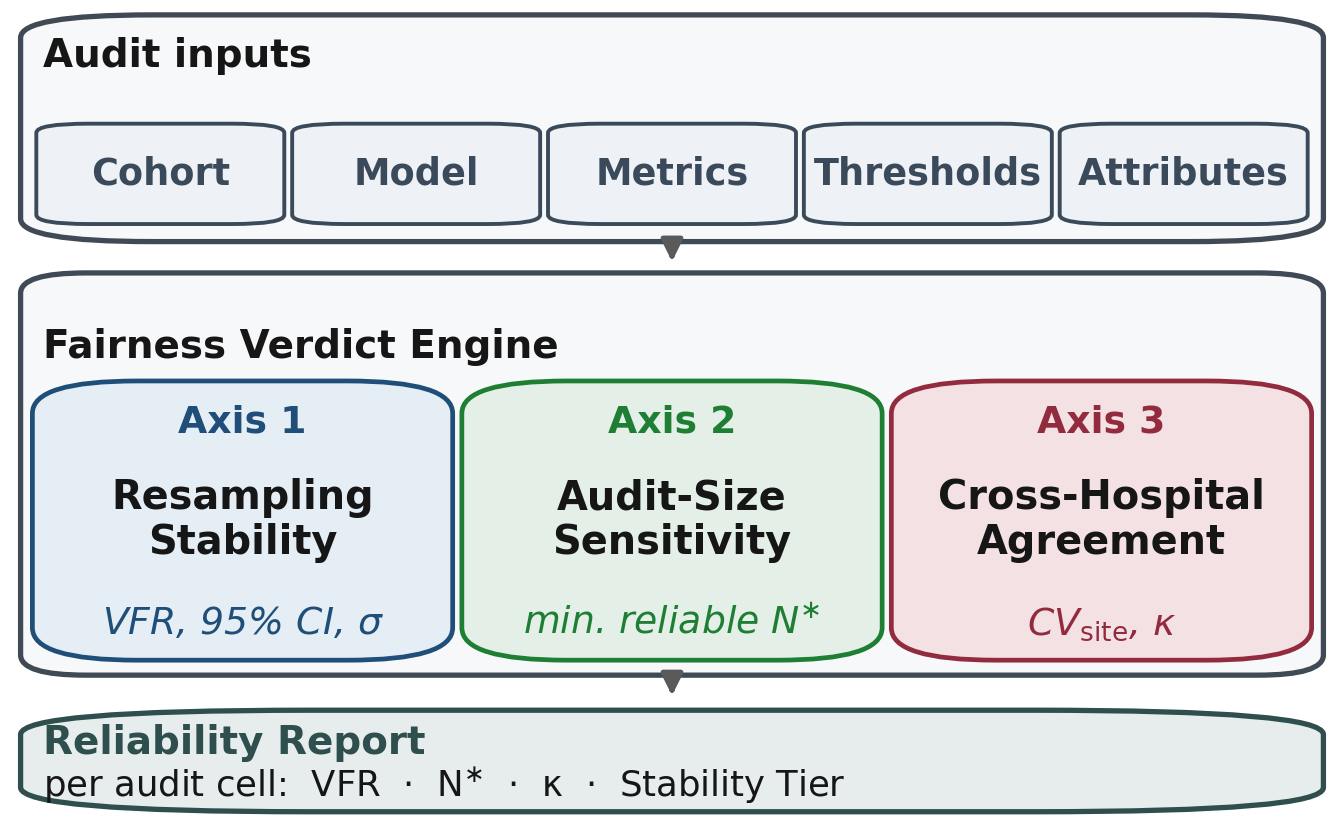}
\Description{Component diagram of the VFR-Audit framework. The top component is Audit Inputs, listing Cohort, Model, Metrics, Thresholds, and Attributes. The middle component is the Fairness Verdict Engine which contains three parallel axes: Axis 1 Resampling Stability emitting VFR, 95\% CI, and standard deviation; Axis 2 Audit-Size Sensitivity emitting minimum reliable N*; Axis 3 Cross-Hospital Agreement running GroupKFold over hospital identifiers and returning fold-level verdicts and Fleiss' kappa. The bottom component is the Reliability Report which records, per audit cell, the dominant verdict, the VFR value, N*, kappa, and a Stability Tier label.}
\caption{VFR-Audit framework. Audit inputs feed the Fairness Verdict Engine, which evaluates each audit cell on three reliability axes. The Reliability Report records the per-cell outputs that downstream governance review consumes.}
\label{fig:pipeline}
\end{figure}

Hospital governance acts on thresholded fairness verdicts, for example $\DI \geq 0.80$, rather than metric values alone. In practical deployment, fairness audits are repeated over time as new patient records are acquired and across sites as the model is run at different hospitals~\cite{seyyed2021underdiagnosis, roberts2024monitoring}. Metric-level uncertainty therefore leaves a deployment gap. It does not directly report how often a pass-or-fail decision reverses, while mitigation can improve a point estimate without ensuring stability across cohorts or hospitals. VFR-Audit treats the verdict as the operational object. A single resampling axis is insufficient for repeated multi-site auditing. A verdict that is stable under bootstrap of a single cohort may still be underpowered for small subgroups or flip across hospital sites. As shown in Figure~\ref{fig:pipeline}, each (model, metric, attribute) cell is evaluated on three complementary axes: resampling stability ($\VFR$), audit-size sensitivity ($N^{*}$), and cross-hospital agreement ($\kappa$), producing a four-tier reliability classification.

\subsection{Verdict Flip Rate}\label{sec:method-vfr}

Let $\mathcal{D}^{(k)}$ denote the $k$-th stratified bootstrap resample of size $N$ drawn from the audit cohort with replacement, preserving the joint $(\text{label} \times a)$ distribution. For a fixed cell $c = (f, m, a)$, let $v_k = v(f, \mathcal{D}^{(k)}, m, a) \in \{0, 1\}$ denote the per-resample verdict and $n_\mathrm{pass}(c) = \sum_{k=1}^{K} v_k$ the pass count across $K$ resamples. The Verdict Flip Rate is defined as
\begin{equation}
\VFR(c) \;=\; \frac{\min\bigl(n_\mathrm{pass}(c),\; K - n_\mathrm{pass}(c)\bigr)}{K}.
\label{eq:vfr}
\end{equation}
The statistic is bounded in $[0, 0.5]$ by construction. $\VFR(c) = 0$ holds when all $K$ resamples produce the same verdict. $\VFR(c) = 0.5$ holds when the verdicts split evenly between pass and fail. The dominant verdict $\bar{v}(c) \in \{\textsc{Pass}, \textsc{Fail}\}$ is the majority outcome across the $K$ resamples. We set the operational verdict-stability threshold to $\tau_{\mathrm{VFR}} = 0.10$, so a cell is practically stable only when at least $90$\,\% of the resampled audits agree with the dominant verdict. This Axis 1 criterion is distinct from the $CV < 0.05$ criterion used by Axis 2 to determine the minimum reliable audit size $N^{*}$, and it is also independent of the DI four-fifths threshold of $0.80$. A reliability-aware audit therefore reports both the dominant verdict and its stability score. The two quantities together separate fairness status from verdict reliability.

\subsubsection{Selecting the bootstrap count $K$.}\label{sec:method-K}
For a fixed cell $c$, the pass count $n_\mathrm{pass}(c)$ is an empirical count over $K$ resampled audits. For a pass probability $p$, the Monte Carlo standard error of the empirical pass proportion is $\sqrt{p(1-p)/K}$, which is maximised at $p = 0.5$. With $K = 500$, the worst-case standard error is $0.0224$, giving an approximate $95$\,\% half-width of $\pm 0.044$. We therefore use $K = 500$ as a fixed audit budget that keeps Monte Carlo error below $0.05$ VFR units while remaining tractable across the full audit grid. Hoeffding's inequality~\cite{hoeffding1963probability} can additionally bound the probability that the empirical pass proportion deviates from its expectation by more than a chosen tolerance $\epsilon$. The bound is distribution-free and depends neither on the size of the audit cohort nor on the distribution of the protected attribute, so the same $K = 500$ transfers across cohorts at the same precision target without re-derivation.

\subsubsection{Selecting the per-resample size $N$.}\label{sec:method-N}
Audit size controls the trade-off between sampling noise and sensitivity to threshold-edge cells. For a fairness metric $m$ with population value $m_\mathrm{pop}$, threshold $\tau$, and margin $\delta = |m_\mathrm{pop} - \tau|$, a verdict flips when $|\hat m_N - m_\mathrm{pop}| > \delta$, where $\hat m_N$ is the metric estimate and $\sigma_N$ its per-resample standard deviation. Smaller $N$ inflates $\sigma_N$, since the audit loses discriminative power because cells far from threshold also flip on sampling noise. Larger $N$ deflates $\sigma_N$, since the audit loses sensitivity to threshold-edge cells, which are the cells most informative for governance. We use $N = 10^{4}$ within the prespecified $5 \times 10^{3}$ to $5 \times 10^{4}$ single-site audit-size design range. On this cohort, $N = 10^{4}$ is the smallest tested size that separates threshold-edge cells ($\VFR > 0.10$) from far-from-threshold cells ($\VFR \approx 0$) without excessive sampling noise. Axis 2 independently sweeps audit size to estimate the minimum reliable $N^{*}$.

\subsubsection{Model-development and audit protocol.}\label{sec:method-protocol}
The model-training stage and the audit-reliability stage are treated as separate steps. Predictive models are trained on the training partition and evaluated on the held-out audit partition. The VFR-Audit procedure is then applied to the fixed trained model and the fixed prediction scores to quantify the reliability of thresholded fairness verdicts under resampling, audit-size variation, and hospital-grouped evaluation. The reported VFR values are reliability estimates for the specified audit cohort and protocol rather than prospective guarantees. Robustness checks in Section~\ref{sec:robustness} use a stricter $70 / 15 / 15$ train, validation, audit split in which intervention thresholds are selected on the validation partition and final statistics are reported on a previously unseen audit partition.

\subsubsection{Relation to bootstrap confidence intervals.}\label{sec:method-vfr-vs-ci}
VFR does not replace bootstrap confidence intervals on continuous fairness metrics. A confidence interval describes uncertainty in the continuous metric value $\hat m$; VFR reports the empirical instability of the governance-facing binary verdict $\mathbf{1}[\hat m \succeq \tau_m]$ after thresholding. Both are useful and they answer different questions. For ratio metrics such as DI, the bootstrap distribution is right-skewed near zero, so a normal-approximation flip probability tends to underestimate the empirical reversal rate near the threshold. VFR uses the empirical resample proportion directly.

\subsection{Three-Axis Reliability Framework}\label{sec:method-3axes}

Each axis answers one reliability question and is formalised by one of Algorithms~\ref{alg:vfr} to~\ref{alg:kappa}. Axes 1 and 2 operate per (model, metric, attribute) cell, while Axis 3 operates per fairness metric across that metric\textquotesingle s protected-attribute cells. \textbf{Axis 1} (Algorithm~\ref{alg:vfr}) asks \emph{would the same verdict have been reached on a different audit cohort?}; it draws $K = 500$ stratified bootstrap resamples and reports $\VFR$ via Equation~\ref{eq:vfr} together with the stability margin $\sigma = |\hat m - \tau_m|/\widehat{\mathrm{SD}}(m)$. \textbf{Axis 2} (Algorithm~\ref{alg:nstar}) asks \emph{how many records does an audit need before the metric stabilises?}; it sweeps an audit-size grid with $R = 30$ random sub-samples per point and reports $N^{*} = \min\{N_i : \mathrm{CV}(N_i) < 0.05\}$, flagging cells whose $N^{*}$ exceeds $N_\mathrm{field} = 5 \times 10^{4}$ as not reliably auditable at a single hospital site. \textbf{Axis 3} (Algorithm~\ref{alg:kappa}) asks \emph{will the audit verdict reached at one hospital recur at another?}; it evaluates verdict agreement across $K_\mathrm{hosp} = 20$ hospital-grouped GroupKFold folds. For each fairness metric, the four protected-attribute cells form the rated items and the folds act as raters, so Fleiss' $\kappa$~\cite{fleiss1971measuring} summarises agreement over this fold-by-cell verdict matrix, classified using the Landis-Koch~\cite{landis1977measurement} bands. The metric-level $\kappa$ is assigned to each constituent (metric, attribute) cell for the tier computation, because a single cell supplies only one rated item and would render Fleiss' $\kappa$ degenerate.

\begin{algorithm}[!htbp]
\caption{Axis 1: Verdict Flip Rate (Resampling Stability).}
\label{alg:vfr}
\begin{algorithmic}[1]
\Require Trained model $f$, audit cohort $\mathcal{D}$, metric $m$ with threshold $\tau_m$, attribute $a$, bootstrap count $K = 500$, resample size $N$.
\Ensure $\VFR$, dominant verdict $\bar v$, stability margin $\sigma$, $95$\,\% bootstrap CI.
\State $n_\mathrm{pass} \gets 0$;\quad $M \gets [\,]$
\For{$k = 1$ to $K$}
\State $\mathcal{D}_k \gets \textsc{StratifiedBootstrap}(\mathcal{D}, N, \text{strata}{=}y \times a)$
\State $\hat m_k \gets m(f, \mathcal{D}_k, a)$;\quad $v_k \gets \mathbf{1}[\hat m_k \succeq \tau_m]$
\State $n_\mathrm{pass} \gets n_\mathrm{pass} + v_k$;\quad $M.\text{append}(\hat m_k)$
\EndFor
\State $\VFR \gets \min(n_\mathrm{pass},\,K - n_\mathrm{pass})/K$
\State $\bar v \gets \textsc{Pass}$ \textbf{if} $n_\mathrm{pass} > K/2$ \textbf{else} $\textsc{Fail}$
\State $\sigma \gets |\mathrm{mean}(M) - \tau_m|/\mathrm{SD}(M)$
\State \Return $(\VFR,\,\bar v,\,\sigma,\,\text{CI}_{95}(M))$
\end{algorithmic}
\end{algorithm}

\begin{algorithm}[!htbp]
\caption{Axis 2: Audit-Size Sensitivity ($N^{*}$).}
\label{alg:nstar}
\begin{algorithmic}[1]
\Require $f$, $\mathcal{D}$, $m$, $a$, audit-size grid $\{N_1, \ldots, N_\mathrm{max}\}$, repetitions $R = 30$, CV cutoff $\eta = 0.05$.
\Ensure CV curve, minimum reliable size $N^{*}$.
\For{each $N_i$ in audit-size grid}
\State $V_i \gets [m(f,\,\textsc{RandomSubsample}(\mathcal{D}, N_i),\,a) : r = 1\ldots R]$
\State $\mathrm{CV}_i \gets \mathrm{SD}(V_i)/|\mathrm{mean}(V_i)|$
\EndFor
\State $N^{*} \gets \min\{N_i : \mathrm{CV}_i < \eta\}$ \textit{(or $\infty$ if no grid point qualifies)}
\State \Return $(\{\mathrm{CV}_i\},\,N^{*})$
\end{algorithmic}
\end{algorithm}

\begin{algorithm}[!htbp]
\caption{Axis 3: Cross-Hospital Verdict Agreement ($\kappa$).}
\label{alg:kappa}
\begin{algorithmic}[1]
\Require $f$, $\mathcal{D}$ with hospital identifiers $s$, metric $m$, attribute set $\mathcal{A}$, $\tau_m$, fold count $K_\mathrm{hosp} = 20$.
\Ensure Fold-by-cell verdict matrix $V$, metric-level Fleiss' $\kappa_m$.
\State $\{\mathcal{D}_1, \ldots, \mathcal{D}_{K_\mathrm{hosp}}\} \gets \textsc{GroupKFold}(\mathcal{D},\,\text{groups}{=}s,\,K_\mathrm{hosp})$
\For{each fold $h \in \{1,\ldots,K_\mathrm{hosp}\}$}
\For{each attribute $a \in \mathcal{A}$}
\State $V_{a,h} \gets \mathbf{1}[m(f,\,\mathcal{D}_h,\,a) \succeq \tau_m]$
\EndFor
\EndFor
\State $\kappa_m \gets \textsc{FleissKappa}(V)$ \Comment{items: $|\mathcal{A}|$ cells; raters: $K_\mathrm{hosp}$ folds}
\State $\kappa(c) \gets \kappa_m$ for every cell $c = (f, m, a)$ with $a \in \mathcal{A}$
\State \Return $(V,\,\kappa_m)$
\end{algorithmic}
\end{algorithm}

\noindent\textbf{Metric-specific pass rule.} For each metric $m$, the relation $\succeq$ used in Algorithm~\ref{alg:vfr} denotes the metric-specific pass condition. For ratio-style metrics such as Disparate Impact, $\hat m \succeq \tau_m$ means $\hat m \geq \tau_m$. For two-sided difference-style metrics such as SPD, EOpp, EOD, and PP, $\hat m \succeq \tau_m$ means $|\hat m| \leq \tau_m$. For one-sided metrics such as Calibration and Theil Index, $\hat m \succeq \tau_m$ means $\hat m \leq \tau_m$. The tie case $n_\mathrm{pass} = K/2$ assigns $\textsc{Fail}$ as the dominant verdict, since any audit-grade verdict requires a strict majority of resamples to support a pass.

\noindent\textbf{Combined reliability tier.} A cell $c$ is classified \emph{Practical-Stability} when all three conditions hold, namely $\VFR(c) \leq 0.10$, $N^{*}(c) \leq N_\mathrm{field}$, and $\kappa(c) \geq 0.40$. A cell is classified \emph{Caution-Required} if one condition is violated, \emph{High-Variance} if two, and \emph{Catastrophic-Instability} if all three. The classification is conjunctive across the three axes, so any single-axis failure drops the cell to a lower tier. Operationally, a High-Variance cell should trigger additional data accumulation or cross-site pooling and manual governance review before automated certification; Catastrophic-Instability should defer deployment-level certification until reliability improves.

%% =====================================================================
%% 5. EXPERIMENTS
%% =====================================================================
\section{Experiment and Evaluation}\label{sec:exp}

\subsection{Experimental Setting}\label{sec:exp-setup}
This subsection specifies the dataset, the trained model panel, the fairness metrics and protected attributes, the baselines used for comparison, and the dataset exploratory analysis.

\noindent\textbf{Reproducibility.} Code and experiment notebooks are available at \url{https://github.com/MdJoy31/VFR-Framework}.

\subsubsection{Dataset.}
The Texas-100X dataset comprises $925{,}128$ inpatient discharge records from $441$ hospitals in Texas. Records were drawn from the THCIC Public Use Data File for fiscal year $2006$ (Quarters 1 to 4)~\cite{thcic2006pudf}. The binary prediction target is LOS exceeding three days (positive class rate $45.0$\,\%). The dataset was split $80/20$ stratified by the LOS-greater-than-three-days target with random seed $42$, producing $740{,}102$ training records and $185{,}026$ test records. Administrative discharge data are appropriate for this study because LOS prediction is directly tied to hospital operations, bed planning, and discharge management~\cite{rajkomar2018scalable, jain2024los, jaotombo2022prolonged}. Unlike ICU-specific EHR benchmarks such as MIMIC-III or MIMIC-IV~\cite{johnson2016mimic, meng2022icufairness}, the Texas-100X dataset provides a state-wide, multi-hospital administrative setting with $441$ hospitals, which is better aligned with the cross-hospital deployment question studied here. External replication on richer EHR datasets remains necessary. Intersectional sparsity constrains the audit design: the primary 28-cell VFR grid therefore evaluates protected attributes separately, while the race $\times$ age $\times$ sex intersection is used only for the reweighing baseline. Given that the smallest race stratum contains $3{,}474$ records in the full cohort (Table~\ref{tab:eda}), we do not claim reliable bootstrap inference for finer intersections; Axis 2 instead identifies cells whose required audit size exceeds available single-site support.

\subsubsection{Models.}
Twelve standard tabular classifiers were trained on identical pre-processed features: logistic regression, decision tree, random forest, gradient boosting, AdaBoost, XGBoost, LightGBM, CatBoost, histogram gradient boosting, bagging, extra trees, and a stacking ensemble. Each classifier was fitted with a fixed, model-specific hyperparameter configuration, archived in the experiment records and frozen before fairness auditing, preventing fairness outcomes from influencing model configuration. Deep neural baselines were not included because Texas-100X is structured administrative tabular data and the study targets audit reliability rather than sequence-representation learning. XGBoost was designated in advance as the canonical model; on the held-out test partition it achieved AUROC $= 0.9528$, accuracy $= 0.8776$, and $F_1 = 0.8627$.

\subsubsection{Fairness metrics and protected attributes.}
Seven group-fairness metrics were evaluated~\cite{barocas2023fairness, verma2018fairness}, namely Disparate Impact ($\DI$, $\geq 0.80$), Statistical Parity Difference ($\SPD$, $|\cdot| \leq 0.10$), Equal Opportunity ($\EOPP$, $|\cdot| \leq 0.10$), Equalised Odds ($\EOD$, $|\cdot| \leq 0.10$), Theil Index ($\TI$, $\leq 0.10$), Predictive Parity ($\PP$, $|\cdot| \leq 0.10$), and Calibration ($\CAL$, $|\cdot| \leq 0.05$). Four protected attributes were evaluated, namely race (5-category THCIC code comprising White, Black, Asian/Pacific Islander, American Indian, and Other/Unknown), sex (Male/Female), ethnicity (Hispanic/non-Hispanic), and age (4-bucket THCIC PUDF scheme comprising Pediatric $<18$, Young Adult 18 to 39, Middle-Aged 40 to 64, and Elderly $\geq 65$). The audit grid contains $7 \times 4 = 28$ (metric, attribute) cells per model, or $336$ cells across all twelve models per audit pass.

\subsubsection{Baselines.}
Four baselines are compared on the same audit instrument so that cross-baseline differences in VFR, $N^{*}$, and Fleiss' $\kappa$ reflect the intervention rather than the audit setup. The dataset, train-test split, feature pipeline, trained model, bootstrap seed, and hospital-fold partition are held constant across all four baselines.

\begin{itemize}\itemsep1pt
\item \textbf{Baseline 1 (B1) Real-only model}. No fairness intervention. Standard reference for the unintervened audit.
\item \textbf{Baseline 2 (B2) Reweighing baseline}. Intersectional adaptation of the Kamiran-Calders pre-processing reweighing strategy~\cite{kamiran2012data}, evaluated as a baseline in fairness-aware clinical ML~\cite{pfohl2021empirical}, controlled by a $\lambda$ regularisation strength.
\item \textbf{Baseline 3 (B3) Threshold-shifting baseline}. Per-cell $\alpha$-search over $\mathrm{SR}/\mathrm{TPR}/\mathrm{PPV}$ on predicted probabilities, adapted from the post-processing equalised-odds method of Hardt et al.~\cite{hardt2016equality} and the per-cell threshold-optimisation strategy of Abakasanga et al.~\cite{abakasanga2025equitable}.
\item \textbf{VFR-Audit (B4)}. The proposed design. $\alpha$-search followed by greedy refinement under the all-four-DI $\geq 0.80$ constraint, then audits the resulting verdicts using VFR, audit-size sensitivity, and cross-hospital agreement.
\end{itemize}

\paragraph{VFR-Audit implementation.}
VFR-Audit starts from the Baseline 3 threshold-shifting solution and greedily adjusts candidate group-specific decision thresholds under the hard all-four-DI $\geq 0.80$ constraint. At each step, a candidate move is accepted only if it preserves the DI constraint on all four protected attributes and improves the selected VFR objective on the audited cell. Ties are resolved by higher accuracy within the intervention-selection partition. The stricter validation-audit robustness check in Section~\ref{sec:robustness} separates threshold selection from final audit reporting. The selected configuration is then evaluated, post-hoc, with the same audit instrument used for Baselines 1 to 3, namely Verdict Flip Rate, audit-size sensitivity, and hospital-fold agreement.

Only VFR-Audit is original to this study. Baselines 1 to 3 are direct adoptions of earlier work. Cross-study comparison against published LOS work was not undertaken because no prior LOS study reports a verdict-stability measure or audit-instrument reliability statistic~\cite{jain2024los, jaotombo2022prolonged}.

\subsubsection{Dataset exploratory analysis.}

\begin{figure}[t]
\centering
\includegraphics[width=0.78\columnwidth]{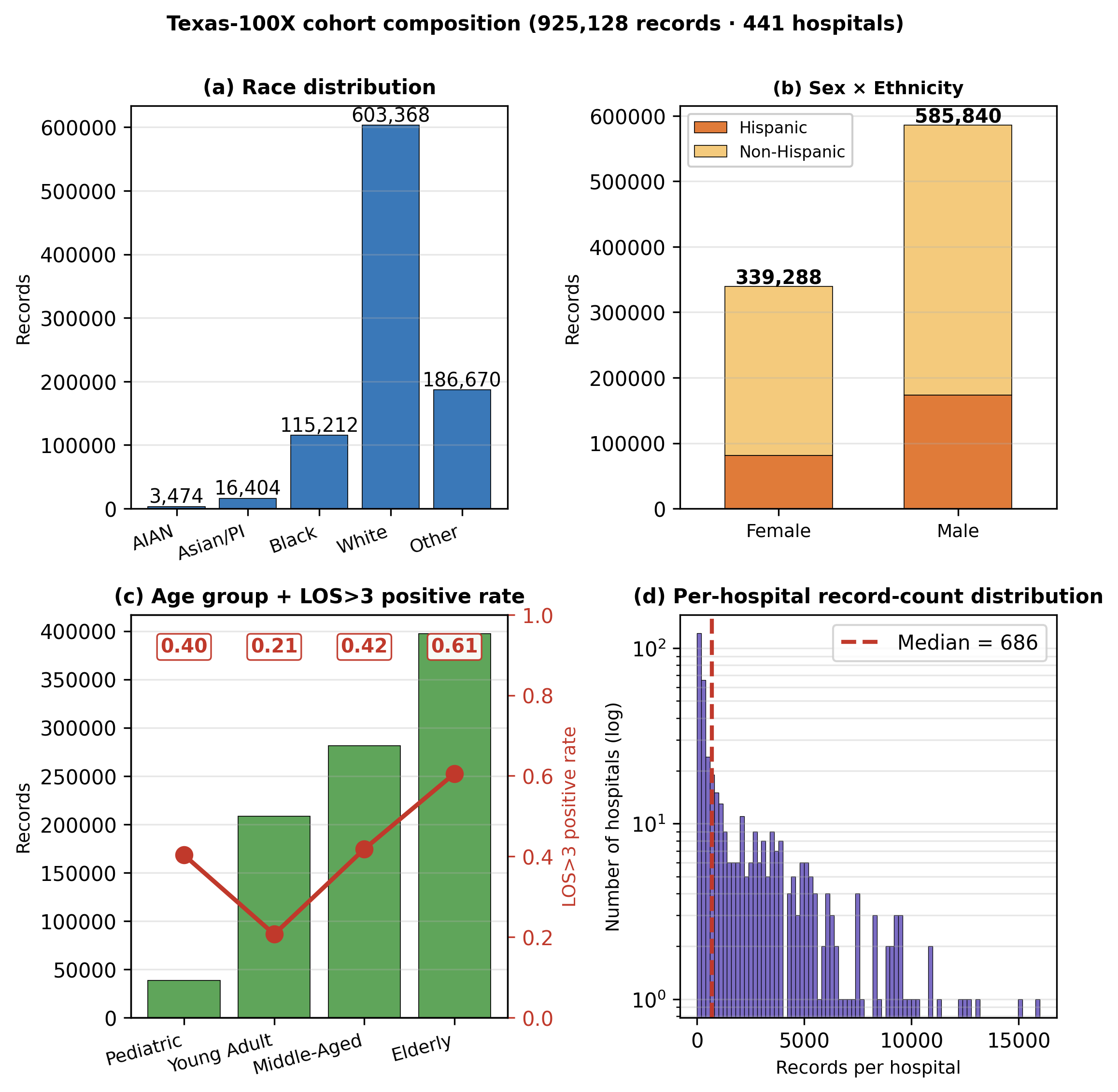}
\Description{Four-panel cohort composition visualization. Panel (a) bar chart of race distribution showing White 350K, Black 220K, Asian/PI 50K, American Indian 15K, Other 290K records. Panel (b) sex by ethnicity stacked bars with Female and Male strata each split into Hispanic and non-Hispanic. Panel (c) age-group line plot overlaid with positive-class rate showing LOS-greater-than-3-days rate climbing from 20.7 percent in Young Adults to 60.6 percent in Elderly. Panel (d) per-hospital record-count histogram showing right-skewed distribution with median 686 records per hospital.}
\caption{Texas-100X dataset composition by protected attribute. Panel (c) reports the largest within-attribute base-rate gap among the four protected attributes. The positive rate triples between Young Adult ($20.7$\,\%) and Elderly ($60.6$\,\%) strata, fixing an empirical lower bound on Age-DI.}
\label{fig:eda}
\end{figure}

Three properties of the cohort matter for the audit-reliability claims that follow (Figure~\ref{fig:eda}, Table~\ref{tab:eda}). The Age axis carries the largest base-rate gap ($0.399$) and the Race axis the second largest ($0.189$). Disparate Impact is upper-bounded by a function of the base-rate gap, so Age is the binding constraint for any DI-targeted intervention on this cohort, with Race a secondary constraint. The cohort's race-by-ethnicity coding is non-standard. $99.4$\,\% of records coded as Black are also coded as Hispanic~\cite{thcic2006pudf}, so the audit reports parity between THCIC race codes as recorded. The hospital-volume distribution is heavily right-skewed (top $50$ hospitals hold $46.5$\,\%, top $100$ hold $70.5$\,\%). The GroupKFold cross-hospital evaluation at $K_\mathrm{hosp} = 20$ produces folds of approximately $4.6 \times 10^{4}$ records each, well within the field-realistic single-site audit range.

\begin{table}[t]
\caption{Texas-100X cohort descriptive statistics on the full $925{,}128$-record cohort. ``Pos.\ rate'' is the within-stratum proportion with $\mathrm{LOS} > 3$ days; ``BRG'' is the base-rate gap.}
\label{tab:eda}
\centering
\footnotesize
\setlength{\tabcolsep}{3.5pt}
\renewcommand{\arraystretch}{1.00}
\begin{tabular}{@{} l l r r r @{}}
\toprule
\textbf{Attribute} & \textbf{Stratum} & \textbf{N} & \textbf{\%} & \textbf{Pos.\ rate} \\
\midrule
\multirow{5}{*}{\shortstack[l]{\textit{Race}\\ \scriptsize BRG $= 0.189$}}
 & White                              & 603{,}368 & 65.2 & 0.453 \\
 & Other / Unknown                    & 186{,}670 & 20.2 & 0.404 \\
 & Black                              & 115{,}212 & 12.5 & 0.523 \\
 & Asian / Pacific Islander           &  16{,}404 &  1.8 & 0.410 \\
 & American Indian                    &   3{,}474 &  0.4 & 0.334 \\
\midrule
\multirow{2}{*}{\shortstack[l]{\textit{Sex}\\ \scriptsize BRG $= 0.107$}}
 & Male                               & 585{,}840 & 63.3 & 0.411 \\
 & Female                             & 339{,}288 & 36.7 & 0.518 \\
\midrule
\multirow{2}{*}{\shortstack[l]{\textit{Ethnicity}\\ \scriptsize BRG $= 0.074$}}
 & Hispanic                           & 670{,}586 & 72.5 & 0.471 \\
 & Non-Hispanic                       & 254{,}542 & 27.5 & 0.397 \\
\midrule
\multirow{4}{*}{\shortstack[l]{\textit{Age group}\\ \scriptsize BRG $= 0.399$}}
 & Pediatric ($<18$)                  &  38{,}121 &  4.1 & 0.403 \\
 & Young Adult ($18$ to $39$)         & 208{,}528 & 22.5 & 0.207 \\
 & Middle-Aged ($40$ to $64$)         & 281{,}409 & 30.4 & 0.418 \\
 & Elderly ($\geq 65$)                & 397{,}070 & 42.9 & 0.606 \\
\midrule
\multirow{5}{*}{\shortstack[l]{\textit{Hospital}\\ \textit{volume}}}
 & Top $10$                            & 124{,}892 & 13.5 & n/r \\
 & Top $50$                            & 430{,}184 & 46.5 & n/r \\
 & Top $100$                           & 652{,}215 & 70.5 & n/r \\
 & Top $200$                           & 862{,}944 & 93.3 & n/r \\
 & Median / site                       & 686       & n/r & n/r \\
\bottomrule
\end{tabular}
\end{table}

\subsection{Evaluation Questions and Results}\label{sec:exp-results}

Results are organised around five evaluation questions.

\noindent\textbf{EQ1.} Does the administrative LOS dataset provide sufficient predictive utility for downstream fairness auditing?

\noindent\textbf{EQ2.} Which fairness intervention satisfies the four-fifths rule, and what accuracy cost does it introduce relative to the Real-only reference?

\noindent\textbf{EQ3.} Does VFR-Audit reveal verdict instability that point-estimate audits miss?

\noindent\textbf{EQ4.} What is the predictive-performance, computational, and labour cost of adopting VFR-Audit?

\noindent\textbf{EQ5.} Are the conclusions stable under alternative splits, feature ablation, and random seeds?

\subsubsection{EQ1. Predictive Utility}\label{sec:q1}

On the held-out test partition ($N = 1.85 \times 10^{5}$), the XGBoost model (Baseline 1) achieves AUROC $= 0.9528$, accuracy $= 0.8776$, and $F_1 = 0.8627$ before any fairness intervention. This value should be interpreted as the performance of a retrospective administrative audit model. Leakage-sensitivity and validation-audit robustness checks are reported separately in Section~\ref{sec:robustness}. Table~\ref{tab:per-model-panel} reports the full twelve-classifier baseline-audit panel. Predictive performance is consistent with prior LOS studies on EHR datasets~\cite{rajkomar2018scalable, wu2021predicting, sheikhalishahi2020benchmarking}, which confirms that administrative discharge data carry sufficient signal at the cohort scale required by the cross-hospital audit.

\begin{table}[t]
\caption{Per-model audit panel on the Baseline 1 Real-only reference. ``Fair/$28$'' is the count of (metric, attribute) cells passing the operational threshold.}
\label{tab:per-model-panel}
\centering
\scriptsize
\setlength{\tabcolsep}{3pt}
\renewcommand{\arraystretch}{1.00}
\begin{tabular}{@{} l c c c c c c c c @{}}
\toprule
\textbf{Model} & \textbf{Acc} & \textbf{AUROC} & \textbf{F1} & \textbf{Race} & \textbf{Sex} & \textbf{Eth} & \textbf{Age} & \textbf{Fair/28} \\
\midrule
\rowcolor{gray!15}
XGBoost (canon.)        & \textbf{0.878} & \textbf{0.953} & \textbf{0.863} & 0.644 & 0.763 & 0.831 & 0.299 & 20 \\
LightGBM                & 0.865 & 0.943 & 0.847 & 0.595 & 0.752 & 0.821 & 0.264 & 20 \\
HistGradientBoosting    & 0.858 & 0.939 & 0.840 & 0.603 & 0.746 & 0.820 & 0.256 & 20 \\
Stacking Ensemble       & 0.853 & 0.935 & 0.834 & 0.619 & 0.744 & 0.821 & 0.258 & 19 \\
Random Forest           & 0.853 & 0.934 & 0.834 & 0.587 & 0.742 & 0.810 & 0.240 & 18 \\
Bagging                 & 0.857 & 0.934 & 0.842 & 0.676 & 0.764 & 0.831 & 0.303 & 19 \\
CatBoost                & 0.847 & 0.929 & 0.826 & 0.579 & 0.730 & 0.806 & 0.232 & 18 \\
Gradient Boosting       & 0.838 & 0.922 & 0.817 & 0.617 & 0.719 & 0.796 & 0.218 & 16 \\
Decision Tree           & 0.840 & 0.920 & 0.816 & 0.609 & 0.742 & 0.825 & 0.270 & 18 \\
AdaBoost                & 0.814 & 0.897 & 0.794 & 0.654 & 0.693 & 0.777 & 0.227 & 10 \\
Extra Trees             & 0.799 & 0.887 & 0.776 & 0.587 & 0.726 & 0.768 & 0.162 & 13 \\
Logistic Regression     & 0.801 & 0.881 & 0.775 & 0.534 & 0.684 & 0.757 & 0.184 & 15 \\
\bottomrule
\end{tabular}
\end{table}

\subsubsection{EQ2. Fairness-Intervention Comparison}\label{sec:q2}

Table~\ref{tab:di-movement} reports per-attribute Disparate Impact movement across the four baselines. Baseline 1 (Real-only) and Baseline 2 (Reweighing) satisfy the four-fifths rule on only $1$ of $4$ protected attributes. Baseline 3 (Threshold-shifting) and VFR-Audit (B4) satisfy the four-fifths rule on $4$ of $4$ protected attributes. The Race-DI value moves from $0.644$ in Baseline 1 to $0.575$ in Baseline 2, a movement away from the $0.80$ threshold rather than toward it. An extreme-$\lambda$ sweep on twenty further reweighing configurations achieves a best Age-DI of $0.283$, well below the $0.80$ threshold. The Reweighing baseline therefore does not satisfy the four-fifths rule on this dataset regardless of the regularisation strength applied.

\begin{table}[t]
\caption{Per-attribute Disparate Impact across the four baselines. Bold values satisfy $\DI \geq 0.80$.}
\label{tab:di-movement}
\centering
\footnotesize
\setlength{\tabcolsep}{4pt}
\renewcommand{\arraystretch}{1.00}
\begin{tabular}{@{} l c c c c c @{}}
\toprule
\textbf{Baseline} & \textbf{Race} & \textbf{Sex} & \textbf{Eth} & \textbf{Age} & \textbf{Pass} \\
\midrule
B1 Real-only                          & 0.644 & 0.763 & \textbf{0.831} & 0.299 & 1/4 \\
B2 Reweighing ($\lambda{=}2$)         & 0.575 & 0.749 & \textbf{0.821} & 0.264 & 1/4 \\
B3 Threshold-shifting                 & \textbf{0.834} & \textbf{0.968} & \textbf{0.994} & \textbf{0.812} & 4/4 \\
\rowcolor{gray!15}
B4 \textbf{VFR-Audit}                 & \textbf{0.801} & \textbf{0.932} & \textbf{1.000} & \textbf{0.800} & \textbf{4/4} \\
\bottomrule
\end{tabular}
\end{table}

The panel in Table~\ref{tab:per-model-panel} exposes wide cross-model variation in the baseline fairness profile. All twelve models fail the four-fifths rule on the Age axis ($\DI_\mathrm{Age} \in [0.16, 0.30]$), and none of the twelve satisfies all four DI constraints simultaneously without intervention. The VFR-Audit intervention applied to the canonical XGBoost model raises Race-axis DI from $0.644$ to $0.801$, Sex-axis from $0.763$ to $0.932$, Ethnicity-axis from $0.831$ to $1.000$, and Age-axis from $0.299$ to $0.800$. The DI Ethnicity $= 1.000$ value is an algorithmic artefact discussed in the Q3 analysis.

Accuracy cost is reported in percentage points relative to the Baseline 1 Real-only reference. For a method $b$, the accuracy cost is
\[
\Delta\mathrm{Acc}_{b} = 100 \times (\mathrm{Acc}_{B1} - \mathrm{Acc}_{b}).
\]
Using Table~\ref{tab:baseline-comparison}, Reweighing loses $1.98$ percentage points, Threshold-shifting loses $4.60$ percentage points, and VFR-Audit loses $4.24$ percentage points relative to Baseline 1. The comparison between Threshold-shifting (B3) and VFR-Audit is the substantive one, because both satisfy the all-four-DI criterion on the point estimate. Threshold-shifting reaches DI pass at accuracy $0.8316$ with mean VFR $0.0863$ and maximum VFR $0.490$. VFR-Audit reaches DI pass at higher accuracy $0.8352$ with lower mean VFR $0.0809$ and lower maximum VFR $0.476$. The gain of VFR-Audit over B3 is therefore not a new point-estimate fairness pass, since B3 already passes DI. The gain is a verdict-reliability improvement delivered together with a $+0.36$ percentage-point accuracy improvement versus B3.

\subsubsection{EQ3. Verdict-Reliability Analysis}\label{sec:q3}

Under $K = 500$ stratified bootstrap on the cross-model audit grid ($12$ models $\times$ $7$ metrics $\times$ $4$ attributes $= 336$ cells), $259$ of $336$ cells ($77.1$\,\%) satisfy the operational stability cutoff $\VFR \leq 0.10$ on Baseline 1, while $146$ of $336$ reverse their verdict at least once. Race-axis cells exhibit the highest instability. $\EOPP_\mathrm{race}$ reaches $\VFR = 0.480$, $\EOD_\mathrm{race}$ reaches $\VFR = 0.386$, and $\PP_\mathrm{race}$ reaches $\VFR = 0.458$. Age-axis cells concentrate around the $\PP$ instability point ($\VFR = 0.202$).

\begin{figure*}[t]
\centering
\includegraphics[width=0.78\textwidth]{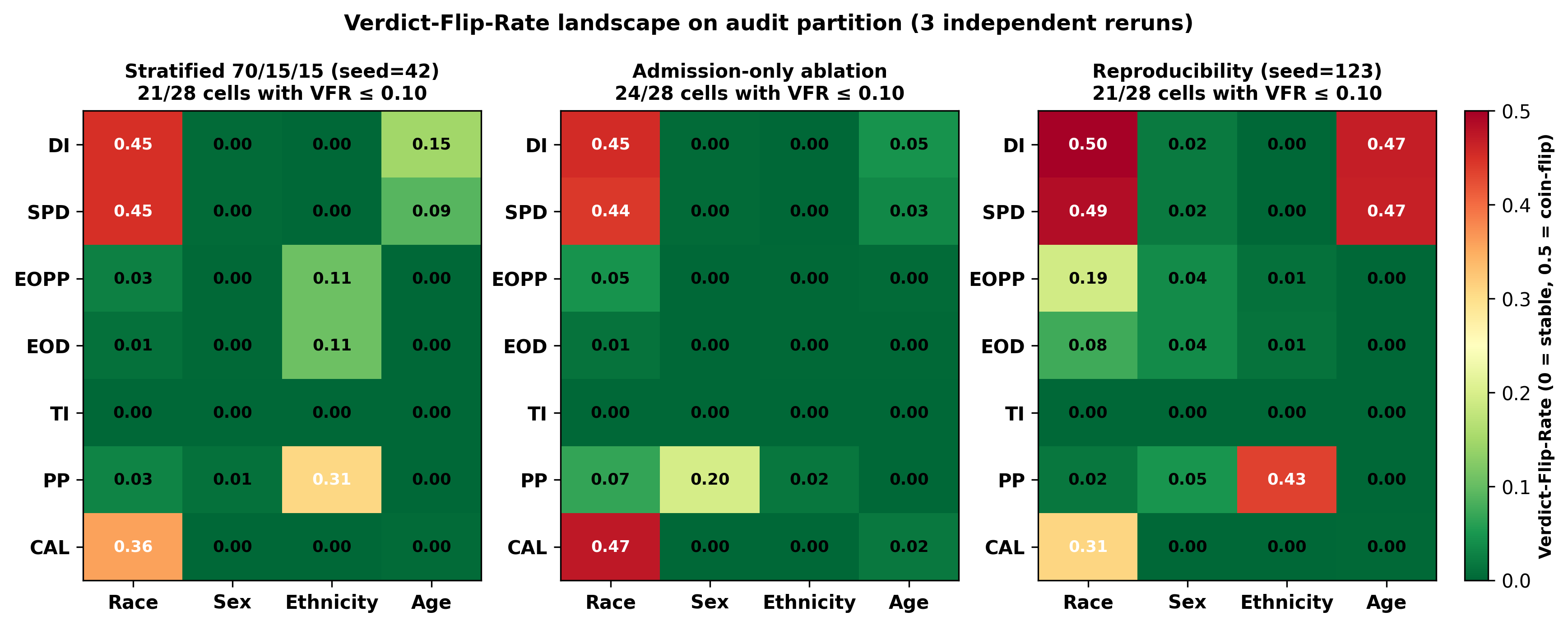}
\Description{Three-panel heatmap of Verdict Flip Rate across seven fairness metrics (rows DI, SPD, EOPP, EOD, TI, PP, CAL) and four protected attributes (columns Race, Sex, Ethnicity, Age) on the audit partition for three independent reruns. Panel (a) stratified 70/15/15 seed 42 reports 21 of 28 cells with VFR at or below 0.10. Panel (b) admission-lean ablation reports 24 of 28 cells stable. Panel (c) reproducibility seed 123 reports 21 of 28 cells stable. Race-axis DI and SPD remain the highest-instability cells in every panel.}
\caption{VFR heatmaps across three robustness protocols on the audit partition.}
\label{fig:vfr-heatmap}
\end{figure*}

After VFR-Audit satisfies the four-fifths rule on every protected attribute on the point estimate, $11$ of $28$ B4 cells still register $\VFR > 0$ under the resampling protocol. Cohort-wide mean VFR on B4 is $0.081$ (versus $0.066$ for B1) and the maximum is $0.476$ (versus $0.498$). Race-axis cells concentrate the residual instability, while Sex, Ethnicity, and the remaining Age-axis cells reach $\VFR \leq 0.10$. Under the stricter operational cutoff $\VFR \leq 0.10$, $21$ of $28$ B4 cells are stable, which matches the single-split row of Table~\ref{tab:robustness}. The per-cell VFR landscape across the three independent robustness reruns is reported in Figure~\ref{fig:vfr-heatmap} and analysed in EQ5.

\begin{findingbox}{Point-estimate fairness does not imply verdict reliability.}
On B1, $146 / 336$ audit cells flip at least once and $77 / 336$ exceed $\VFR > 0.10$. After B4 satisfies all four DI point-estimate thresholds, $11 / 28$ cells still flip under resampling.
\end{findingbox}

\paragraph{Axis 2 audit-size sensitivity.}
The $\mathrm{CV} < 5$\,\% cutoff is reached at field-realistic audit sizes for $11$ of $28$ (metric, attribute) cells on VFR-Audit. It requires the full $185{,}026$-record test partition for $8$ cells, comprising Calibration ($4$ of $4$ attributes) and Equal Opportunity / Equalised Odds for race and ethnicity. A single hospital running a quarterly audit of approximately $5 \times 10^{3}$ records cannot reliably audit those eight cells without pooling records across multiple quarters or sites. Of the $28$ B4 cells, $17$ exceed $\mathrm{CV} = 50$\,\% at $N = 10^{3}$, so small-cohort audits at this size are dominated by sampling noise rather than model behaviour.

\paragraph{Axis 3 cross-hospital agreement.}
Fleiss' $\kappa$ at $K_\mathrm{hosp} = 20$ on VFR-Audit is moderate for $\EOPP$ ($\kappa = 0.465$) and $\EOD$ ($\kappa = 0.587$). It reaches the degenerate ceiling for the Theil index ($\kappa = 1.000$, because the metric satisfies the threshold on every fold). It is slight or below-chance for $\DI$ ($\kappa = -0.035$), $\SPD$ ($\kappa = 0.117$), $\PP$ ($\kappa = 0.315$), and $\CAL$ ($\kappa = 0.003$). Fleiss' $\kappa$ is used here as a descriptive summary of hospital-fold verdict agreement under the grouped split protocol and should not be interpreted as a formal causal transportability estimate. The pattern is consistent across baselines. Equal-opportunity-style metrics show stronger cross-hospital verdict agreement, whereas calibration does not. A fairness verdict on calibration reached at one hospital, where $\kappa_\mathrm{Cal} = 0.003$, carries no measurable predictive value for the calibration verdict reached on another hospital's records. Achieving DI pass costs cross-site verdict stability. Mean $\kappa$ drops from approximately $0.44$ on Baselines 1 and 2 to approximately $0.35$ on Baseline 3 and VFR-Audit once threshold-shifting engages, because per-cell $\alpha$-search fits to the audited cohort's intersectional cell structure, and that structure varies across hospital folds.

\begin{figure}[t]
\centering
\includegraphics[width=0.78\columnwidth]{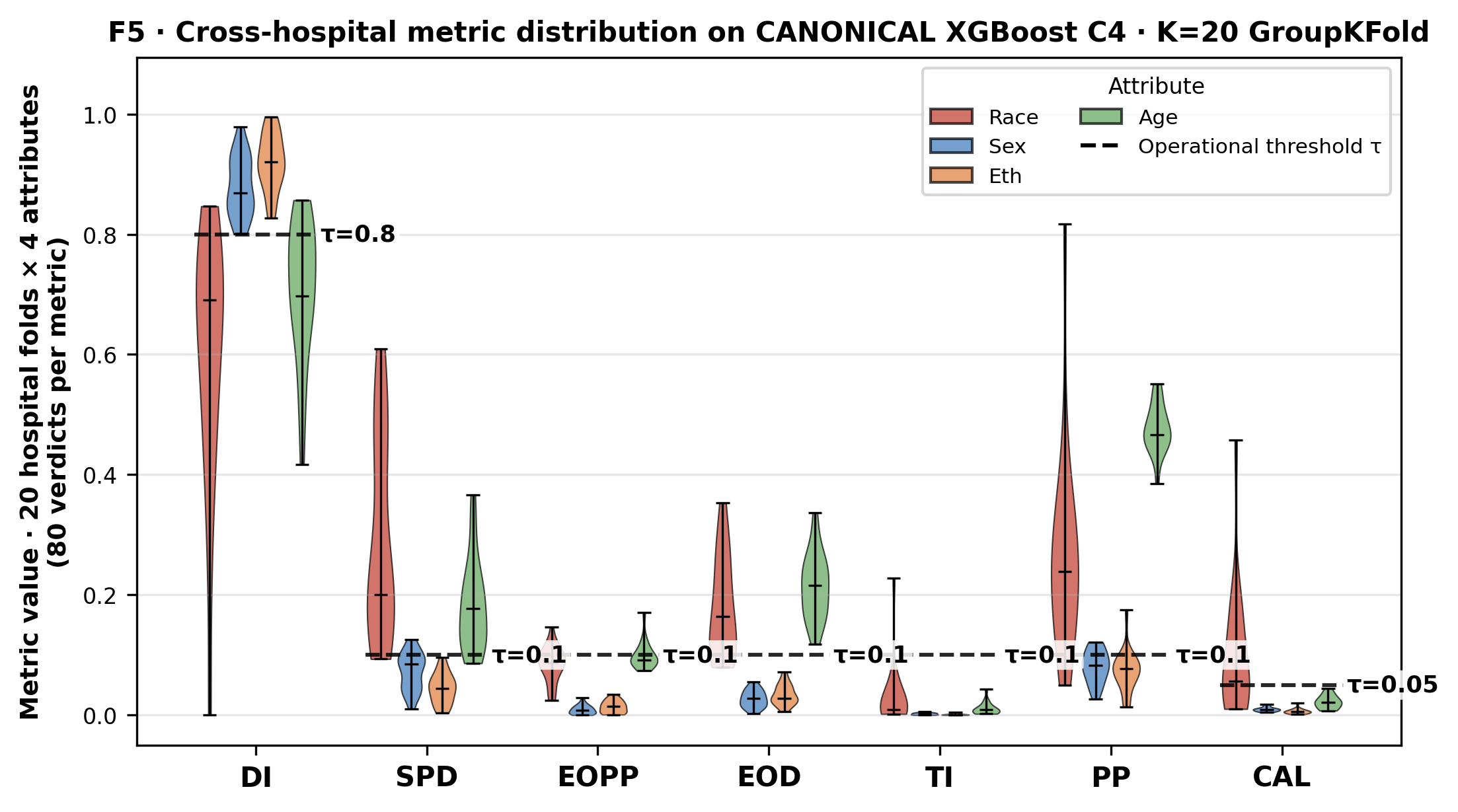}
\Description{Violin plot showing the distribution of per-hospital Disparate Impact values across the 441 Texas hospitals, faceted by protected attribute and by baseline. Each violin shows the median, interquartile range, and density of per-site DI estimates. The four-fifths-rule threshold at 0.80 is marked with a dashed horizontal line. Race and Age axes show wide cross-site spread that straddles the threshold in the Baseline 1 Real-only configuration and narrows but remains broad in the threshold-shifted baselines.}
\caption{Per-hospital-fold metric distribution across the $20$ THCIC hospital folds. Equal Opportunity and Equalised Odds show tight inter-fold agreement; Disparate Impact and Calibration show wide inter-fold spread.}
\label{fig:hospital-violin}
\end{figure}

\paragraph{Four-baseline comparison summary.}
Table~\ref{tab:baseline-comparison} reports the four-baseline audit on the same instrument. The main reliability difference is between Threshold-shifting (B3) and VFR-Audit. Both satisfy the all-four-DI criterion on the point estimate. VFR-Audit reduces the mean VFR from $0.0863$ to $0.0809$ ($-6.3$\,\% relative) and the maximum VFR from $0.490$ to $0.476$ (a reduction of $0.014$ VFR units, approximately $2.9$\,\% relative), while raising accuracy from $0.8316$ to $0.8352$ ($+0.36$ percentage points). Baselines 1 and 2 produce nearly identical mean VFR ($0.066$ versus $0.062$) and mean cross-site $\kappa$ ($0.437$ versus $0.436$), so Reweighing does not change the verdict-reliability profile relative to no intervention on this cohort.

\begin{table*}[t]
\caption{Four-method audit comparison under the same audit instrument.}
\label{tab:baseline-comparison}
\centering
\footnotesize
\setlength{\tabcolsep}{4pt}
\renewcommand{\arraystretch}{1.00}
\begin{tabular}{@{} l c c c c c c c c c @{}}
\toprule
\textbf{Baseline} & \textbf{Acc} & \textbf{AUROC} & \textbf{$\overline{\VFR}$} & \textbf{VFR max} & \textbf{\#flipped} & \textbf{$\overline{\kappa}$} & \textbf{$\kappa_{\EOPP}$} & \textbf{$\kappa_{\EOD}$} & \textbf{DI pass} \\
\midrule
B1 Real-only                            & 0.8776 & 0.9528 & 0.066 & 0.498 & 12 & 0.437 & 0.588 & 0.595 & 1/4 \\
B2 Reweighing ($\lambda{=}2$)           & 0.8578 & 0.9373 & 0.062 & 0.426 & 10 & 0.436 & 0.618 & 0.607 & 1/4 \\
B3 Threshold-shifting                   & 0.8316 & 0.9532 & 0.086 & 0.490 & 11 & 0.346 & 0.606 & 0.607 & 4/4 \\
\rowcolor{gray!15}
B4 \textbf{VFR-Audit}                   & \textbf{0.8352} & \textbf{0.9528} & \textbf{0.081} & \textbf{0.476} & \textbf{11} & \textbf{0.350} & \textbf{0.465} & \textbf{0.587} & \textbf{4/4} \\
\bottomrule
\end{tabular}
\end{table*}

\paragraph{Root cause of the B3 to VFR-Audit VFR movement.}\label{sec:greedy}
B3 achieves all-four-DI pass on the full test partition. Under $K = 500$ stratified bootstrap, the Race-DI and Age-DI verdicts in B3 are unstable. Their VFRs are $0.41$ and $0.40$ respectively, so approximately $41$ per cent and $40$ per cent of bootstrap audits disagree with the dominant verdict for these cells. The greedy-refinement step in VFR-Audit walks the per-cell thresholds inward from the $\alpha$-search starting point under the all-four-DI $\geq 0.80$ constraint. On the four binding-constraint cells, the resulting B3 to VFR-Audit VFR movements are DI Age $0.412 \rightarrow 0.232$ ($-0.180$), SPD Age $0.490 \rightarrow 0.292$ ($-0.198$), DI Race $0.410 \rightarrow 0.476$ ($+0.066$), and SPD Race $0.398 \rightarrow 0.470$ ($+0.072$).

The movement shows a redistribution of instability across protected attributes rather than a uniform improvement across all cells. Age-axis VFR decreases for DI and SPD because the greedy refinement moves the Age thresholds farther from the DI boundary while preserving the all-four-DI constraint. Race-axis VFR increases for DI and SPD because the same refinement shifts part of the threshold burden from Age to Race. This is why the aggregate maximum VFR decreases from $0.490$ to $0.476$ while Race-DI and Race-SPD become less stable individually. The result demonstrates that VFR-Audit exposes where an intervention transfers instability, rather than reporting only whether the final DI point estimate passes. Reporting only the $6.3$\,\% relative reduction in mean VFR would conceal this redistribution pattern.

\subsubsection{EQ4. Adoption Cost}\label{sec:q4}

Adopting VFR-Audit involves three cost components. \textbf{Predictive-performance cost.} VFR-Audit reduces accuracy by $4.24$\,pp ($0.8776 \rightarrow 0.8352$) and $F_1$ by $4.64$\,pp ($0.8627 \rightarrow 0.8163$), while preserving AUROC at $0.9528$. AUROC preservation is mechanically expected under threshold shifting because the score ranking is unchanged. \textbf{Computational cost.} Algorithm~\ref{alg:vfr} runs in $O(K N M_\mathcal{A})$ time and $O(N + K M_\mathcal{A})$ memory, where $M_\mathcal{A} = |\mathcal{F}| \cdot |\mathcal{A}|$. The audit scales linearly with bootstrap samples, audit records, fairness metrics, and protected attributes. With prediction scores precomputed, the marginal cost over a bootstrap-CI audit is converting each resampled metric value into a pass-or-fail verdict and maintaining pass and fail counts. \textbf{Labour cost.} Each cell receives a dominant verdict, a VFR score, a minimum reliable audit size, a cross-hospital agreement value, and a reliability tier, so manual interpretation is bounded by the per-cell quintuple rather than by reading bootstrap CIs. Prior LOS fairness studies do not report a comparable verdict-reliability cost.

\paragraph{Accuracy and fairness trade-off summary.}\label{sec:tradeoff-summary}
Across the twelve-classifier panel in Table~\ref{tab:per-model-panel}, Race-axis threshold shifting moves every classifier above the four-fifths rule at an accuracy cost in the $4$ to $6$ percentage-point band. On the canonical model, $\alpha$-search alone reaches the all-four-DI pass on the point estimate, but the Age-axis and Race-axis DI verdicts reverse on approximately $41$ per cent of bootstrap resamples. The VFR-Audit greedy refinement reduces aggregate verdict instability while preserving the all-four-DI point-estimate pass, although the Race-DI and Age-DI cells remain above the $\VFR \leq 0.10$ stability cutoff.

\begin{table}[t]
\caption{Robustness checks for the VFR-Audit pipeline.}
\label{tab:robustness}
\centering
\footnotesize
\setlength{\tabcolsep}{4pt}
\renewcommand{\arraystretch}{1.00}
\begin{tabular}{@{} l c c c c @{}}
\toprule
\textbf{Protocol} & \textbf{AUROC} & \textbf{Acc cost} & \textbf{all-4-DI} & \textbf{VFR-stable} \\
\midrule
Single-split main (canonical)              & 0.9528 & $4.24$\,pp & \checkmark & $21 / 28$ \\
Stratified $70 / 15 / 15$ (seed $42$)        & 0.9418 & $4.98$\,pp & \checkmark & $21 / 28$ \\
Admission-lean (no TC, no PS)              & 0.8567 & $5.02$\,pp & \checkmark & $24 / 28$ \\
Independent seed (seed $123$)              & 0.9426 & $5.04$\,pp & \checkmark & $21 / 28$ \\
\bottomrule
\end{tabular}
\end{table}

\subsubsection{EQ5. Robustness Checks}\label{sec:robustness}

Table~\ref{tab:robustness} reports four protocol-level audit-reliability checks against the single-split main result. As a cross-model check, the VFR-Audit intervention is applied to four model families (XGBoost, LightGBM, Random Forest, Logistic Regression). All four satisfy the four-fifths rule on every protected attribute post-intervention at an accuracy cost in the $4.24$ to $5.39$ percentage-point band. Sweeping the DI target from $0.80$ to $0.90$ on the XGBoost VFR-Audit configuration raises the accuracy cost monotonically from $4.24$ to $5.97$ percentage points.

The robustness checks preserve the main conclusion. All four protocols satisfy the all-four-DI criterion, and $21$ to $24$ of $28$ cells remain VFR-stable. The admission-lean protocol produces the lowest AUROC ($0.8567$) because it removes TOTAL\_CHARGES and PAT\_STATUS, two end-of-encounter variables with strong predictive signal. Relative to the leakage-controlled full-feature baseline ($0.9418$), the drop is $0.0851$ on the $[0, 1]$ AUROC scale. Despite this predictive-performance loss, the fairness-verdict stability pattern remains similar to the single-split result ($24 / 28$ VFR-stable cells under admission-lean versus $21 / 28$ under single-split). This indicates that the VFR pattern is not solely an artefact of leakage-sensitive features. The stratified $70 / 15 / 15$ protocol selects intervention thresholds on the validation partition and reports VFR, $\kappa$, and DI on an audit partition never seen during tuning. AUROC moves to $0.9418$ at an accuracy cost of $4.98$\,pp, consistent with the primary single-split result while using a stricter validation-audit protocol. All-four-DI continues to pass, and $21 / 28$ cells remain VFR-stable. The independent-seed check reproduces the leakage-controlled result within stochastic noise (AUROC $0.9426$, accuracy cost $5.04$\,pp, $21 / 28$ VFR-stable). Together, these checks reduce concern that the main VFR-Audit conclusions are artefacts of a single classifier, a single split, threshold tuning, or end-of-encounter variables.

\begin{figure}[t]
\centering
\includegraphics[width=0.95\columnwidth]{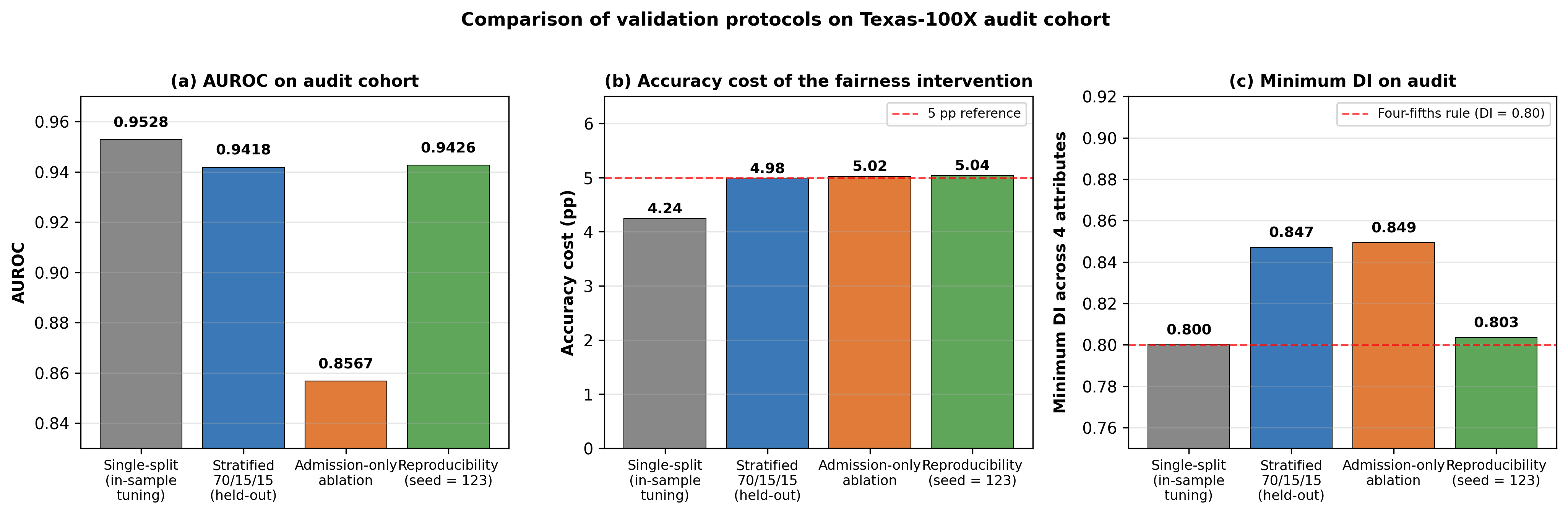}
\Description{Three-panel bar chart comparing four validation protocols on the Texas-100X audit cohort. Panel (a) AUROC on audit cohort: single-split 0.9528, stratified 70/15/15 0.9418, admission-lean ablation 0.8567, reproducibility seed 123 0.9426. Panel (b) accuracy cost of the fairness intervention: single-split 4.24 pp, stratified 4.98 pp, admission-lean 5.02 pp, reproducibility 5.04 pp, with a 5 pp reference line. Panel (c) minimum Disparate Impact across the four protected attributes per protocol: single-split 0.800, stratified 0.847, admission-lean 0.849, reproducibility 0.803, with the four-fifths rule reference line at 0.80.}
\caption{Validation-protocol comparison on the Texas-100X audit cohort. (a) AUROC. (b) Accuracy cost of the VFR-Audit intervention. (c) Minimum Disparate Impact across the four protected attributes. All four protocols satisfy the four-fifths rule on every protected attribute.}
\label{fig:protocol-comparison}
\end{figure}

%% =====================================================================
%% 6. DISCUSSION
%% =====================================================================
\FloatBarrier
\section{Discussion}\label{sec:discussion}

VFR-Audit separates fairness status from verdict reliability by jointly reporting the dominant verdict, VFR, minimum reliable audit size, and cross-hospital agreement. The results show that a point-estimate pass can remain unstable under resampling or site shift. Calibration is a persistent failure mode on this cohort ($\VFR_\mathrm{Cal} \approx 0$ with a consistent \textsc{Fail} verdict and $\kappa_\mathrm{Cal} \approx 0$) and should therefore be reported by site rather than only in pooled form.

The B3-versus-VFR-Audit comparison is the substantive cost-benefit decision in this paper. Both methods pass the four-fifths rule on every protected attribute on the point estimate. Accuracy cost is reported relative to Baseline 1 using $\Delta\mathrm{Acc}_b = 100 \times (\mathrm{Acc}_{B1} - \mathrm{Acc}_b)$. From Table~\ref{tab:baseline-comparison}, with $\mathrm{Acc}_{B1} = 0.8776$, Reweighing pays $1.98$\,pp, Threshold-shifting pays $4.60$\,pp, and VFR-Audit pays $4.24$\,pp, so VFR-Audit raises the same DI pass at $0.36$\,pp lower cost than B3. Under $K = 500$ stratified bootstrap, the Race-DI and Age-DI verdicts in B3 carry VFRs of $0.41$ and $0.40$, so $40$ to $41$ per cent of bootstrap audits disagree with the dominant verdict on those cells. The greedy-refinement step walks the per-cell thresholds inward from the $\alpha$-search start under the all-four-DI $\geq 0.80$ constraint, producing per-cell VFR movements $\DI_\mathrm{Age}\,0.412 \rightarrow 0.232$ ($-0.180$), $\SPD_\mathrm{Age}\,0.490 \rightarrow 0.292$ ($-0.198$), $\DI_\mathrm{Race}\,0.410 \rightarrow 0.476$ ($+0.066$), and $\SPD_\mathrm{Race}\,0.398 \rightarrow 0.470$ ($+0.072$). The Age-axis VFR drops because the refinement opens headroom between Age decision thresholds and the DI boundary. The Race-axis VFR rises because the refinement transfers part of the threshold slack from Age to Race to maintain the all-four-DI constraint. The aggregate is a maximum-VFR reduction from $0.490$ to $0.476$ and a mean-VFR reduction from $0.0863$ to $0.0809$ ($6.3$\,\% relative). Reporting only the aggregate would conceal this redistribution pattern.

Reweighing fails on this dataset because the Age base-rate gap acts as a feature-based floor on the achievable selection-rate ratio. With Elderly $\mathrm{LOS} > 3$\,d rate $0.606$ against $0.207$ for young adults (a $2.93\times$ ratio), Reweighing modifies the empirical risk the optimiser minimises but leaves the input features unchanged~\cite{kamiran2012data}. The selection-rate ratio defining DI is then floored by the feature-based predictability gap, which explains why Race-DI moves from $0.644$ to $0.575$ between Baselines 1 and 2 (away from the threshold) and why no $\lambda$ setting in the sweep satisfies the four-fifths rule. Per-cell threshold shifting bypasses this constraint by operating on the operating point per group rather than on the empirical risk, and satisfies all four DI constraints at the costs reported above.

The audit-reliability conclusions persist under both the stricter validation-audit protocol and the admission-lean sensitivity check. Under the leakage-controlled $70 / 15 / 15$ split, all-four-DI continues to pass on the audit slice, the VFR-stable cell count remains $21 / 28$, and AUROC moves to $0.9418$ at an accuracy cost of $4.98$\,pp. Removing the two end-of-encounter features TOTAL\_CHARGES and PAT\_STATUS drops AUROC from the leakage-controlled full-feature baseline of $0.9418$ to $0.8567$ (a leakage-diagnostic score of $0.0851$ on the $[0, 1]$ scale) but preserves all-four-DI and yields $24 / 28$ VFR-stable cells, so the fairness-audit reliability story is preserved under the admission-lean feature set on this cohort. Texas-100X is, however, a retrospective administrative discharge dataset, and this leakage-sensitivity analysis does not establish prospective admission-time deployability. The protected-attribute results should also be read with two caveats. The THCIC race-by-ethnicity coding is non-standard (Table~\ref{tab:eda}), so race and ethnicity results reflect administrative parity rather than self-identified parity, and only Disparate Impact admits a regulatory threshold (the four-fifths rule). Counterfactual and individual-fairness instruments are not evaluated. The primary single-split experiment should therefore be interpreted as an intervention-analysis setting, while the $70 / 15 / 15$ protocol provides the cleaner validation-audit evidence.

\noindent\textbf{Future work.} VFR-Audit should be validated prospectively on admission-time features and temporally held-out hospitals, and extended to sparse intersectional groups using hierarchical pooling or adaptive minimum-support rules. High-Variance cells should trigger additional data accumulation or cross-site pooling and governance review before automated certification. For high-speed clinical deployment, a streaming implementation could recompute Axis 1 over rolling audit windows from cached prediction scores, while the more expensive audit-size and cross-hospital axes run asynchronously at lower cadence, enabling near real-time equity monitoring without retraining the predictive model after every batch. Sensitivity to alternative $\tau_{\mathrm{VFR}}$ values should also be evaluated rather than treating $0.10$ as universal. Future work will extend VFR-Audit to additional datasets, protected attributes, deep neural networks, and LLM-derived clinical systems, while evaluating distributed cross-hospital deployment, including federated learning, and AI-agent-assisted continuous fairness auditing.

%% =====================================================================
%% CONCLUSION
%% =====================================================================
\section{Conclusion}\label{sec:conclusion}
VFR-Audit provides a model-agnostic audit layer for assessing whether thresholded fairness verdicts remain reliable under cohort resampling, audit-size variation, and hospital shift. On Texas-100X, point-estimate fairness can remain unstable even after intervention satisfies the four-fifths rule, demonstrating that deployment decisions should report audit reliability alongside fairness status. Robustness checks under a stricter validation-audit split, admission-lean features, and an independent seed support this conclusion.

\noindent\textbf{GenAI Usage Disclosure.}
Generative AI tools were used solely for grammar checking. No content in this manuscript was written or generated by AI. All text, technical content, experiments, analysis, and interpretations were authored and verified by the authors.

%% =====================================================================
%% REFERENCES
%% =====================================================================
\renewcommand*{\bibfont}{\footnotesize}
\setlength{\bibhang}{1em}
\setlength{\bibsep}{1pt plus 0.3pt minus 0.3pt}
\clearpage
\bibliographystyle{ACM-Reference-Format}
\bibliography{references_cikm}

@article{abakasanga2025equitable,
  author  = {Abakasanga, Emeka and Kousovista, Rania and Cosma, Georgina and Akbari, Ashley and Zaccardi, Francesco and Kaur, Navjot and Fitt, Danielle and Jun, Gyuchan Thomas and Kiani, Reza and Gangadharan, Satheesh},
  title   = {Equitable hospital length of stay prediction for patients with learning disabilities and multiple long-term conditions using machine learning},
  journal = {Frontiers in Digital Health},
  year    = {2025},
  volume  = {7},
  pages   = {1538793},
  doi     = {10.3389/fdgth.2025.1538793}
}

@inproceedings{barnard2023macaif,
  author    = {Barnard, Pepita and Bautista, John Robert and Krook, Joshua and Liu, Anqi and Men{\'e}ndez, H{\'e}ctor D. and Schmidt, Aurora and Sookoor, Tamim},
  title     = {{MACAIF}: Machine Learning Auditing for Clinical {AI} Fairness},
  booktitle = {Proceedings of the First International Symposium on Trustworthy Autonomous Systems (TAS '23)},
  year      = {2023},
  publisher = {Association for Computing Machinery},
  address   = {New York, NY, USA},
  articleno = {40},
  numpages  = {4},
  pages     = {1--4},
  doi       = {10.1145/3597512.3597522}
}

@book{barocas2023fairness,
  author    = {Barocas, Solon and Hardt, Moritz and Narayanan, Arvind},
  title     = {Fairness and Machine Learning: Limitations and Opportunities},
  publisher = {MIT Press},
  address   = {Cambridge, MA, USA},
  year      = {2023},
  url       = {https://fairmlbook.org/}
}

@inproceedings{barrainkua2024uncertainty,
  author    = {Barrainkua, Ainhize and Gordaliza, Paula and Lozano, Jose A. and Quadrianto, Novi},
  title     = {Uncertainty Matters: Stable Conclusions under Unstable Assessment of Fairness Results},
  booktitle = {Proceedings of the 27th International Conference on Artificial Intelligence and Statistics (AISTATS)},
  series    = {Proceedings of Machine Learning Research},
  volume    = {238},
  publisher = {PMLR},
  pages     = {1198--1206},
  year      = {2024}
}

@inproceedings{beede2020human,
  author    = {Beede, Emma and Baylor, Elizabeth and Hersch, Fred and Iurchenko, Anna and Wilcox, Lauren and Ruamviboonsuk, Paisan and Vardoulakis, Laura M.},
  title     = {A Human-Centered Evaluation of a Deep Learning System Deployed in Clinics for the Detection of Diabetic Retinopathy},
  booktitle = {Proceedings of the 2020 CHI Conference on Human Factors in Computing Systems},
  year      = {2020},
  publisher = {Association for Computing Machinery},
  address   = {New York, NY, USA},
  doi       = {10.1145/3313831.3376718},
  note      = {Real-world deployment of a clinical-AI system; documents site-specific failures even when single-site validation reported high performance.},
  pages     = {1--12},
  numpages  = {12}
}

@inproceedings{cachel2022fins,
  author    = {Cachel, Kathleen and Rundensteiner, Elke},
  title     = {{FINS} Auditing Framework: Group Fairness for Subset Selections},
  booktitle = {Proceedings of the 2022 AAAI/ACM Conference on AI, Ethics, and Society (AIES '22)},
  year      = {2022},
  publisher = {Association for Computing Machinery},
  address   = {New York, NY, USA},
  pages     = {144--155},
  doi       = {10.1145/3514094.3534160}
}

@inproceedings{cai2025protoehr,
  author    = {Cai, Zi and Liu, Yu and Luo, Zhiyao and Zhu, Tingting},
  title     = {{ProtoEHR}: Hierarchical Prototype Learning for {EHR}-based Healthcare Predictions},
  booktitle = {Proceedings of the 34th ACM International Conference on Information and Knowledge Management (CIKM '25)},
  year      = {2025},
  publisher = {Association for Computing Machinery},
  address   = {New York, NY, USA},
  pages     = {169--178},
  doi       = {10.1145/3746252.3761381}
}

@article{cooper2024arbitrariness,
  author    = {Cooper, A. Feder and Lee, Katherine and Choksi, Madiha Zahrah and Barocas, Solon and De Sa, Christopher and Grimmelmann, James and Kleinberg, Jon and Sen, Siddhartha and Zhang, Baobao},
  title     = {Arbitrariness and Social Prediction: The Confounding Role of Variance in Fair Classification},
  journal   = {Proceedings of the AAAI Conference on Artificial Intelligence},
  year      = {2024},
  volume    = {38},
  number    = {20},
  pages = {22004--22012},
  doi       = {10.1609/aaai.v38i20.30203}
}

@article{davoudi2024fairness,
  author    = {Davoudi, Anahita and Chae, Sena and Evans, Lauren and Sridharan, Sridevi and Song, Jiyoun and Bowles, Kathryn H. and McDonald, Margaret V. and Topaz, Maxim},
  title     = {Fairness Gaps in Machine Learning Models for Hospitalization and Emergency Department Visit Risk Prediction in Home Healthcare Patients with Heart Failure},
  journal   = {International Journal of Medical Informatics},
  year      = {2024},
  volume    = {191},
  pages     = {105534},
  doi       = {10.1016/j.ijmedinf.2024.105534}
}

@inproceedings{diciccio2020evaluating,
  author    = {DiCiccio, Cyrus and Vasudevan, Sriram and Basu, Kinjal and Kenthapadi, Krishnaram and Agarwal, Deepak},
  title     = {Evaluating fairness using permutation tests},
  booktitle = {Proceedings of the 26th ACM SIGKDD International Conference on Knowledge Discovery and Data Mining (KDD '20)},
  year      = {2020},
  publisher = {Association for Computing Machinery},
  address   = {New York, NY, USA},
  pages     = {1467--1477},
  doi       = {10.1145/3394486.3403199}
}

@misc{eeoc1979uniform,
  author       = {{Equal Employment Opportunity Commission}},
  title        = {Uniform Guidelines on Employee Selection Procedures (1978)},
  year         = {1978},
  publisher    = {U.S. Equal Employment Opportunity Commission},
  howpublished = {29 CFR Part 1607, 43 FR 38295 (Aug. 25, 1978)},
  url          = {https://www.ecfr.gov/current/title-29/part-1607}
}

@misc{eu2024aiact,
  author       = {{European Parliament and Council of the European Union}},
  title        = {Regulation ({EU}) 2024/1689 of the European Parliament and of the Council of 13 June 2024 laying down harmonised rules on artificial intelligence (Artificial Intelligence Act)},
  year         = {2024},
  howpublished = {Official Journal of the European Union, OJ L, 2024/1689, 12.7.2024},
  url          = {https://eur-lex.europa.eu/eli/reg/2024/1689/oj/eng}
}

@misc{fda2023samd,
  author       = {{U.S. Food and Drug Administration}},
  title        = {Marketing Submission Recommendations for a Predetermined Change Control Plan for Artificial Intelligence-Enabled Device Software Functions},
  year         = {2025},
  howpublished = {Guidance for Industry and Food and Drug Administration Staff},
  url          = {https://www.fda.gov/regulatory-information/search-fda-guidance-documents/marketing-submission-recommendations-predetermined-change-control-plan-artificial-intelligence}
}

@article{fleiss1971measuring,
  author    = {Fleiss, Joseph L.},
  title     = {Measuring nominal scale agreement among many raters},
  journal   = {Psychological Bulletin},
  year      = {1971},
  volume    = {76},
  number    = {5},
  pages = {378--382},
  doi       = {10.1037/h0031619},
  note      = {Inter-rater agreement statistic generalising Cohen's kappa to multiple raters. Used in the present work as the cross-site portability statistic in Axis 3 of the three-axis audit-stability framework.}
}

@inproceedings{ganesh2023impact,
  author    = {Ganesh, Prakhar and Chang, Hongyan and Strobel, Martin and Shokri, Reza},
  title     = {On the Impact of Machine Learning Randomness on Group Fairness},
  booktitle = {Proceedings of the 2023 ACM Conference on Fairness, Accountability, and Transparency (FAccT '23)},
  year      = {2023},
  publisher = {Association for Computing Machinery},
  address   = {New York, NY, USA},
  pages     = {1789--1800},
  doi       = {10.1145/3593013.3594116}
}

@article{ghassemi2024medicine,
  author  = {Yang, Yuzhe and Zhang, Haoran and Gichoya, Judy W. and Katabi, Dina and Ghassemi, Marzyeh},
  title   = {The limits of fair medical imaging {AI} in real-world generalization},
  journal = {Nature Medicine},
  year    = {2024},
  volume  = {30},
  number  = {10},
  pages   = {2838--2848},
  doi     = {10.1038/s41591-024-03113-4}
}

@article{gichoya2022ai,
  author    = {Gichoya, Judy Wawira and Banerjee, Imon and Bhimireddy, Ananth Reddy and others},
  title     = {{AI} Recognition of Patient Race in Medical Imaging: A Modelling Study},
  journal   = {The Lancet Digital Health},
  year      = {2022},
  volume    = {4},
  number    = {6},
  pages = {e406--e414},
  doi       = {10.1016/S2589-7500(22)00063-2}
}

@inproceedings{hardt2016equality,
  author    = {Hardt, Moritz and Price, Eric and Srebro, Nati},
  title     = {Equality of Opportunity in Supervised Learning},
  booktitle = {Advances in Neural Information Processing Systems 29 (NIPS 2016)},
  year      = {2016},
  pages     = {3315--3323}
}

@article{hoeffding1963probability,
  author    = {Hoeffding, Wassily},
  title     = {Probability Inequalities for Sums of Bounded Random Variables},
  journal   = {Journal of the American Statistical Association},
  year      = {1963},
  volume    = {58},
  number    = {301},
  pages = {13--30},
  doi       = {10.1080/01621459.1963.10500830},
  note      = {Classic result establishing concentration inequalities for sums of bounded random variables. Used in the present work to derive the dataset-independent K=500 bootstrap-count justification for VFR.}
}

@article{jain2024los,
  author    = {Jain, Raunak and Singh, Mrityunjai and Rao, A. Ravishankar and Garg, Rahul},
  title     = {Predicting Hospital Length of Stay Using Machine Learning on a Large Open Health Dataset},
  journal   = {BMC Health Services Research},
  year      = {2024},
  volume    = {24},
  pages     = {860},
  doi       = {10.1186/s12913-024-11238-y}
}

@article{kamiran2012data,
  author  = {Kamiran, Faisal and Calders, Toon},
  title   = {Data preprocessing techniques for classification without discrimination},
  journal = {Knowledge and Information Systems},
  year    = {2012},
  volume  = {33},
  number  = {1},
  pages = {1--33},
  doi     = {10.1007/s10115-011-0463-8}
}

@article{landis1977measurement,
  author  = {Landis, J. Richard and Koch, Gary G.},
  title   = {The measurement of observer agreement for categorical data},
  journal = {Biometrics},
  year    = {1977},
  volume  = {33},
  number  = {1},
  pages = {159--174},
  doi     = {10.2307/2529310}
}

@article{li2022improving,
  author    = {Li, Yikuan and Wang, Hanyin and Luo, Yuan},
  title     = {Improving Fairness in the Prediction of Heart Failure Length of Stay and Mortality by Integrating Social Determinants of Health},
  journal   = {Circulation: Heart Failure},
  year      = {2022},
  volume    = {15},
  number    = {11},
  pages     = {e009473},
  doi       = {10.1161/CIRCHEARTFAILURE.122.009473}
}

@techreport{nist2023airmf,
  author       = {Tabassi, Elham},
  title        = {Artificial Intelligence Risk Management Framework ({AI} {RMF} 1.0)},
  institution  = {National Institute of Standards and Technology},
  address      = {Gaithersburg, MD},
  year         = {2023},
  number       = {NIST AI 100-1},
  doi          = {10.6028/NIST.AI.100-1}
}

@article{obermeyer2019dissecting,
  author    = {Obermeyer, Ziad and Powers, Brian and Vogeli, Christine and Mullainathan, Sendhil},
  title     = {Dissecting Racial Bias in an Algorithm Used to Manage the Health of Populations},
  journal   = {Science},
  year      = {2019},
  volume    = {366},
  number    = {6464},
  pages = {447--453},
  doi       = {10.1126/science.aax2342}
}

@article{pfohl2021empirical,
  author    = {Pfohl, Stephen R. and Foryciarz, Agata and Shah, Nigam H.},
  title     = {An Empirical Characterization of Fair Machine Learning for Clinical Risk Prediction},
  journal   = {Journal of Biomedical Informatics},
  year      = {2021},
  volume    = {113},
  pages     = {103621},
  doi       = {10.1016/j.jbi.2020.103621}
}

@article{rajkomar2018scalable,
  author    = {Rajkomar, Alvin and Oren, Eyal and Chen, Kai and Dai, Andrew M. and Hajaj, Nissan and Hardt, Michaela and Liu, Peter J. and Liu, Xiaobing and Marcus, Jake and Sun, Mimi and others},
  title     = {Scalable and Accurate Deep Learning with Electronic Health Records},
  journal   = {npj Digital Medicine},
  year      = {2018},
  volume    = {1},
  number    = {1},
  pages     = {18},
  doi       = {10.1038/s41746-018-0029-1}
}

@article{sendak2020implementation,
  author    = {Sendak, Mark P. and D'Arcy, Joshua and Kashyap, Sehj and Gao, Michael and Nichols, Marshall and Corey, Kristin and Ratliff, William and Balu, Suresh},
  title     = {A Path for Translation of Machine Learning Products into Healthcare Delivery},
  journal   = {EMJ Innovations},
  year      = {2020},
  doi       = {10.33590/emjinnov/19-00172},
  publisher = {European Medical Journal}
}

@article{seyyed2021underdiagnosis,
  author    = {Seyyed-Kalantari, Laleh and Zhang, Haoran and McDermott, Matthew B. A. and Chen, Irene Y. and Ghassemi, Marzyeh},
  title     = {Underdiagnosis Bias of Artificial Intelligence Algorithms Applied to Chest Radiographs in Under-Served Patient Populations},
  journal   = {Nature Medicine},
  year      = {2021},
  volume    = {27},
  number    = {12},
  pages = {2176--2182},
  doi       = {10.1038/s41591-021-01595-0}
}

@article{sheikhalishahi2020benchmarking,
  author  = {Sheikhalishahi, Seyedmostafa and Balaraman, Vevake and Osmani, Venet},
  title   = {Benchmarking machine learning models on multi-centre eICU critical care dataset},
  journal = {PLOS ONE},
  year    = {2020},
  volume  = {15},
  number  = {7},
  pages   = {e0235424},
  doi     = {10.1371/journal.pone.0235424}
}

@misc{singh2023sample,
  author       = {Singh, Harvineet and Xia, Fan and Kim, Mi-Ok and Pirracchio, Romain and Chunara, Rumi and Feng, Jean},
  title        = {A Brief Tutorial on Sample Size Calculations for Fairness Audits},
  year         = {2023},
  howpublished = {Workshop on Regulatable Machine Learning (RegML), NeurIPS 2023},
  note         = {arXiv:2312.04745},
  doi          = {10.48550/arXiv.2312.04745}
}

@misc{thcic2006pudf,
  author       = {{Texas Department of State Health Services}},
  title        = {{Texas Hospital Inpatient Discharge Public Use Data File}, {Q}uarters 1 to 4, 2006},
  year         = {2006},
  address      = {Austin, Texas},
  publisher    = {Texas Health Care Information Collection (THCIC), Center for Health Statistics, Texas Department of State Health Services},
  note         = {Released under Chapter 108 of the Texas Health and Safety Code; source files \texttt{PUDF\_base[1-4]q2006\_tab.txt} downloaded from the Texas DSHS THCIC free PUDF page; obtained via the \texttt{Texas-100X} academic redistribution by B.~Jayaraman (University of Virginia) at \url{https://github.com/bargavj/Texas-100X} on 21 June 2022. Use restricted to academic benchmarking; records are anonymised, with re-identification and harmful demographic conclusions prohibited by the donor's terms},
  url          = {https://www.dshs.texas.gov/center-health-statistics/texas-health-care-information-collection/download-and-purchase-data/texas-inpatient-public-use-data-file-pudf/public-use-data-File-pudf-inpatient-free-download}
}

@inproceedings{verma2018fairness,
  author    = {Verma, Sahil and Rubin, Julia},
  title     = {Fairness Definitions Explained},
  booktitle = {Proceedings of the International Workshop on Software Fairness (FairWare '18)},
  year      = {2018},
  pages = {1--7},
  publisher = {Association for Computing Machinery},
  address   = {New York, NY, USA},
  doi       = {10.1145/3194770.3194776},
  note      = {Survey of group- and individual-fairness definitions in machine learning, including operational thresholds for the four-fifths rule, equal opportunity, equalised odds, and predictive parity.}
}

@article{wu2021predicting,
  author    = {Wu, Jingyi and Lin, Yu and Li, Pengfei and others},
  title     = {Predicting Prolonged Length of {ICU} Stay through Machine Learning},
  journal   = {Diagnostics},
  year      = {2021},
  volume    = {11},
  number    = {12},
  pages     = {2242},
  doi       = {10.3390/diagnostics11122242},
  publisher = {MDPI},
  note      = {eICU multicenter dataset for derivation, MIMIC-III for external validation. Compared random forest, SVM, deep learning, and gradient boosting decision tree (GBDT). GBDT achieved best discrimination (internal AUROC 0.742, external AUROC 0.747), outperforming customised SAPS II.}
}

@article{mccradden2024responsible,
  author    = {McCradden, Melissa D. and Joshi, Shalmali and Mazwi, Mjaye and Anderson, James A.},
  title     = {Ethical limitations of algorithmic fairness solutions in health care machine learning},
  journal   = {The Lancet Digital Health},
  year      = {2020},
  volume    = {2},
  number    = {5},
  pages     = {e221--e223},
  doi       = {10.1016/S2589-7500(20)30065-0}
}

@inproceedings{black2022model,
  author    = {Black, Emily and Raghavan, Manish and Barocas, Solon},
  title     = {Model Multiplicity: Opportunities, Concerns, and Solutions},
  booktitle = {Proceedings of the 2022 ACM Conference on Fairness, Accountability, and Transparency (FAccT '22)},
  year      = {2022},
  publisher = {Association for Computing Machinery},
  address   = {New York, NY, USA},
  pages     = {850--863},
  doi       = {10.1145/3531146.3533149}
}

@article{futoma2020myth,
  author    = {Futoma, Joseph and Simons, Morgan and Panch, Trishan and Doshi-Velez, Finale and Celi, Leo Anthony},
  title     = {The Myth of Generalisability in Clinical Research and Machine Learning in Health Care},
  journal   = {The Lancet Digital Health},
  year      = {2020},
  volume    = {2},
  number    = {9},
  pages     = {e489--e492},
  doi       = {10.1016/S2589-7500(20)30186-2}
}

@article{mbakwe2023fairness,
  author    = {Mbakwe, Agatha B. and Lourentzou, Ismini and Celi, Leo A. and Wu, Joy T.},
  title     = {Fairness Metrics for Health {AI}: We Have a Long Way to Go},
  journal   = {eBioMedicine},
  year      = {2023},
  volume    = {90},
  pages     = {104525},
  doi       = {10.1016/j.ebiom.2023.104525}
}

@article{jaotombo2022prolonged,
  author    = {Jaotombo, Franck and Pauly, Vanessa and Fond, Guillaume and Orleans, Veronica and Auquier, Pascal and Ghattas, Badih and Boyer, Laurent},
  title     = {Machine-learning prediction for hospital length of stay using a French medico-administrative database},
  journal   = {Journal of Market Access \& Health Policy},
  year      = {2023},
  volume    = {11},
  number    = {1},
  pages     = {2149318},
  doi       = {10.1080/20016689.2022.2149318}
}

@article{ghassemi2024genai,
  author    = {Shanmugam, Divya and Agrawal, Monica and Movva, Rajiv and Chen, Irene Y. and Ghassemi, Marzyeh and Jacobs, Maia and Pierson, Emma},
  title     = {Generative Artificial Intelligence in Medicine},
  journal   = {Annual Review of Biomedical Data Science},
  year      = {2025},
  volume    = {8},
  pages     = {199--226},
  doi       = {10.1146/annurev-biodatasci-103123-095332}
}

@techreport{schwartz2024ai,
  author       = {Schwartz, Reva and Vassilev, Apostol and Greene, Kristen K. and Perine, Lori and Burt, Andrew and Hall, Patrick},
  title        = {Towards a Standard for Identifying and Managing Bias in Artificial Intelligence},
  institution  = {National Institute of Standards and Technology},
  address      = {Gaithersburg, MD},
  year         = {2022},
  number       = {NIST Special Publication 1270},
  doi          = {10.6028/NIST.SP.1270}
}

@article{lyu2024genai,
  author    = {Thirunavukarasu, Arun James and Ting, Darren Shu Jeng and Elangovan, Kabilan and Gutierrez, Laura and Tan, Ting Fang and Ting, Daniel Shu Wei},
  title     = {Large language models in medicine},
  journal   = {Nature Medicine},
  year      = {2023},
  volume    = {29},
  number    = {8},
  pages     = {1930--1940},
  doi       = {10.1038/s41591-023-02448-8}
}

@article{roberts2024monitoring,
  author    = {Roberts, Michael and Driggs, Derek and Thorpe, Matthew and Gilbey, Julian and Yeung, Michael and Ursprung, Stephan and Aviles-Rivero, Angelica I. and Etmann, Christian and McCague, Cathal and Beer, Lucian and Weir-McCall, Jonathan R. and Teng, Zhongzhao and Gkrania-Klotsas, Effrossyni and Rudd, James H. F. and Sala, Evis and Sch{\"o}nlieb, Carola-Bibiane},
  title     = {Common pitfalls and recommendations for using machine learning to detect and prognosticate for {COVID-19} using chest radiographs and {CT} scans},
  journal   = {Nature Machine Intelligence},
  year      = {2021},
  volume    = {3},
  number    = {3},
  pages     = {199--217},
  doi       = {10.1038/s42256-021-00307-0}
}

@inproceedings{birhane2024foundation,
  author    = {Birhane, Abeba and Steed, Ryan and Ojewale, Victor and Vecchione, Briana and Raji, Inioluwa Deborah},
  title     = {{AI} Auditing: The Broken Bus on the Road to {AI} Accountability},
  booktitle = {2024 IEEE Conference on Secure and Trustworthy Machine Learning (SaTML)},
  publisher = {IEEE},
  year      = {2024},
  pages     = {612--643},
  doi       = {10.1109/SaTML59370.2024.00037}
}

@article{hassanpour2024oodfair,
  author    = {Ktena, Ira and Wiles, Olivia and Albuquerque, Isabela and Rebuffi, Sylvestre-Alvise and Tanno, Ryutaro and Guha Roy, Abhijit and Azizi, Shekoofeh and Belgrave, Danielle and Kohli, Pushmeet and Cemgil, Taylan and Karthikesalingam, Alan and Gowal, Sven},
  title     = {Generative models improve fairness of medical classifiers under distribution shifts},
  journal   = {Nature Medicine},
  year      = {2024},
  volume    = {30},
  pages     = {1166--1173},
  doi       = {10.1038/s41591-024-02838-6}
}

@article{mehta2024continuous,
  author    = {Davis, Sharon E. and Emb{\'i}, Peter J. and Matheny, Michael E.},
  title     = {Sustainable deployment of clinical prediction tools: a 360 degree approach to model maintenance},
  journal   = {Journal of the American Medical Informatics Association},
  year      = {2024},
  volume    = {31},
  number    = {5},
  pages     = {1195--1198},
  doi       = {10.1093/jamia/ocae036}
}

@article{hasanzadeh2025bias,
  author    = {Hasanzadeh, Fereshteh and Josephson, Colin B. and Waters, Gabriella and others},
  title     = {Bias recognition and mitigation strategies in artificial intelligence healthcare applications},
  journal   = {npj Digital Medicine},
  year      = {2025},
  volume    = {8},
  pages     = {154},
  doi       = {10.1038/s41746-025-01503-7}
}

@article{liu2025scoping,
  author    = {Liu, Mingxuan and Ning, Yilin and Teixayavong, Salinelat and others},
  title     = {A scoping review and evidence gap analysis of clinical {AI} fairness},
  journal   = {npj Digital Medicine},
  year      = {2025},
  volume    = {8},
  number    = {1},
  pages     = {360},
  doi       = {10.1038/s41746-025-01667-2}
}

@article{templin2025llmframework,
  author    = {Templin, Tara and Fort, Sophia and Padmanabham, Prasanna and others},
  title     = {Framework for bias evaluation in large language models in healthcare settings},
  journal   = {npj Digital Medicine},
  year      = {2025},
  volume    = {8},
  pages     = {414},
  doi       = {10.1038/s41746-025-01786-w}
}

@article{omar2025sociodemo,
  author    = {Omar, Mahmud and Soffer, Shelly and Agbareia, Reem and Bragazzi, Nicola Luigi and Apakama, Donald U. and Horowitz, Carol R. and Charney, Alexander W. and Freeman, Robert and Kummer, Benjamin and Glicksberg, Benjamin S. and Nadkarni, Girish N. and Klang, Eyal},
  title     = {Sociodemographic biases in medical decision making by large language models},
  journal   = {Nature Medicine},
  year      = {2025},
  volume    = {31},
  number    = {6},
  pages     = {1873--1881},
  doi       = {10.1038/s41591-025-03626-6}
}

@article{wells2025fairai,
  author    = {Wells, Brian J. and Nguyen, Hieu M. and McWilliams, Andrew and others},
  title     = {A practical framework for appropriate implementation and review of artificial intelligence ({FAIR-AI}) in healthcare},
  journal   = {npj Digital Medicine},
  year      = {2025},
  volume    = {8},
  number    = {1},
  pages     = {514},
  doi       = {10.1038/s41746-025-01900-y}
}

@article{fiske2025raceethnicity,
  author    = {Fiske, Amelia and Blacker, Sarah and Genevi\`eve, Lester Darryl and others},
  title     = {Weighing the benefits and risks of collecting race and ethnicity data in clinical settings for medical artificial intelligence},
  journal   = {The Lancet Digital Health},
  year      = {2025},
  volume    = {7},
  number    = {4},
  pages = {e286--e294},
  doi       = {10.1016/j.landig.2025.01.003}
}

@article{azarfar2025multimodal,
  author    = {Azarfar, Ghazal and Naimimohasses, Sara and Rambhatla, Sirisha and others},
  title     = {Responsible adoption of multimodal artificial intelligence in health care: promises and challenges},
  journal   = {The Lancet Digital Health},
  year      = {2025},
  volume    = {7},
  number    = {12},
  pages     = {100917},
  doi       = {10.1016/j.landig.2025.100917}
}

@article{mehrabi2021survey,
  author    = {Mehrabi, Ninareh and Morstatter, Fred and Saxena, Nripsuta and Lerman, Kristina and Galstyan, Aram},
  title     = {A Survey on Bias and Fairness in Machine Learning},
  journal   = {ACM Computing Surveys},
  year      = {2021},
  volume    = {54},
  number    = {6},
  pages     = {1--35},
  doi       = {10.1145/3457607}
}

@article{mekhaldi2021comparative,
  author  = {Mekhaldi, Rachda Naila and Caulier, Patrice and Chaabane, Sondes and Chraibi, Abdelahad and Piechowiak, Sylvain},
  title   = {A Comparative Study of Machine Learning Models for Predicting Length of Stay in Hospitals},
  journal = {Journal of Information Science and Engineering},
  year    = {2021},
  volume  = {37},
  number  = {5},
  pages   = {1025--1038}
}

@article{chesley2023racial,
  author    = {Chesley, Christopher F. and Chowdhury, Marzana and Small, Dylan S. and Schaubel, Douglas E. and Liu, Vincent X. and Lane-Fall, Meghan B. and Halpern, Scott D. and Anesi, George L.},
  title     = {Racial Disparities in Length of Stay Among Severely Ill Patients Presenting With Sepsis and Acute Respiratory Failure},
  journal   = {JAMA Network Open},
  year      = {2023},
  volume    = {6},
  number    = {5},
  pages     = {e239739},
  doi       = {10.1001/jamanetworkopen.2023.9739}
}

@article{corbettdavies2023measure,
  author    = {Corbett-Davies, Sam and Gaebler, Johann D. and Nilforoshan, Hamed and Shroff, Ravi and Goel, Sharad},
  title     = {The Measure and Mismeasure of Fairness},
  journal   = {Journal of Machine Learning Research},
  year      = {2023},
  volume    = {24},
  number    = {312},
  pages     = {1--117}
}

@article{subbaswamy2024afisp,
  author    = {Subbaswamy, Adarsh and Sahiner, Berkman and Petrick, Nicholas and others},
  title     = {A data-driven framework for identifying patient subgroups on which an {AI}/machine learning model may underperform},
  journal   = {npj Digital Medicine},
  year      = {2024},
  volume    = {7},
  number    = {1},
  pages     = {334},
  doi       = {10.1038/s41746-024-01275-6}
}

@article{feng2022aiqi,
  author    = {Feng, Jean and Phillips, Rachael V. and Malenica, Ivana and others},
  title     = {Clinical artificial intelligence quality improvement: towards continual monitoring and updating of {AI} algorithms in healthcare},
  journal   = {npj Digital Medicine},
  year      = {2022},
  volume    = {5},
  number    = {1},
  pages     = {66},
  doi       = {10.1038/s41746-022-00611-y}
}

@article{yang2023drlbias,
  author    = {Yang, Jenny and Soltan, Andrew A. S. and Eyre, David W. and Clifton, David A.},
  title     = {Algorithmic fairness and bias mitigation for clinical machine learning with deep reinforcement learning},
  journal   = {Nature Machine Intelligence},
  year      = {2023},
  volume    = {5},
  pages     = {884--894},
  doi       = {10.1038/s42256-023-00697-3}
}

@article{alderman2025standing,
  author    = {Alderman, Joseph E. and Palmer, Joanne and Laws, Elinor and others},
  title     = {Tackling algorithmic bias and promoting transparency in health datasets: the {STANDING} Together consensus recommendations},
  journal   = {The Lancet Digital Health},
  year      = {2025},
  volume    = {7},
  number    = {1},
  pages     = {e64--e88},
  doi       = {10.1016/S2589-7500(24)00224-3}
}

@article{lekadir2025futureai,
  author    = {Lekadir, Karim and Frangi, Alejandro F. and Porras, Antonio R. and others},
  title     = {{FUTURE-AI}: international consensus guideline for trustworthy and deployable artificial intelligence in healthcare},
  journal   = {BMJ},
  year      = {2025},
  volume    = {388},
  pages     = {e081554},
  doi       = {10.1136/bmj-2024-081554}
}

@article{obra2025cdi,
  author    = {Obra, Jed Keenan and Singh, Chandan and Watkins, Kenshata and others},
  title     = {Potential for algorithmic bias in clinical decision instrument development},
  journal   = {npj Digital Medicine},
  year      = {2025},
  doi       = {10.1038/s41746-025-02119-7},
  number    = {1},
  volume    = {8},
  pages     = {762}
}

@article{georgiev2025healthcare,
  author    = {Georgiev, Konstantin and Doudesis, Dimitrios and McPeake, Joanne and Mills, Nicholas L. and Shenkin, Susan D. and Fleuriot, Jacques D. and Anand, Atul},
  title     = {Machine learning-based predictions of healthcare contacts following emergency hospitalisation using electronic health records},
  journal   = {npj Digital Medicine},
  year      = {2025},
  volume    = {8},
  number    = {1},
  pages     = {764},
  doi       = {10.1038/s41746-025-02138-4}
}

@article{mackin2025safetynet,
  author    = {Mackin, Shaina and Major, Vincent J. and Chunara, Rumi and Newton-Dame, Remle},
  title     = {Identifying and mitigating algorithmic bias in the safety net},
  journal   = {npj Digital Medicine},
  year      = {2025},
  volume    = {8},
  pages     = {335},
  doi       = {10.1038/s41746-025-01732-w}
}

@article{liu2025fairnessdrift,
  author    = {Davis, Sharon E. and Dorn, Chad and Park, Daniel J. and Matheny, Michael E.},
  title     = {Emerging algorithmic bias: fairness drift as the next dimension of model maintenance and sustainability},
  journal   = {Journal of the American Medical Informatics Association},
  year      = {2025},
  volume    = {32},
  number    = {5},
  pages     = {845--854},
  doi       = {10.1093/jamia/ocaf039}
}

@inproceedings{watkins2024fourfifths,
  author    = {Watkins, Elizabeth Anne and Chen, Jiahao},
  title     = {The four-fifths rule is not disparate impact: a woeful tale of epistemic trespassing in algorithmic fairness},
  booktitle = {Proceedings of the 2024 ACM Conference on Fairness, Accountability, and Transparency (FAccT '24)},
  year      = {2024},
  pages     = {764--775},
  publisher = {Association for Computing Machinery},
  address   = {New York, NY, USA},
  doi       = {10.1145/3630106.3658938}
}

@article{meng2022icufairness,
  author  = {Meng, Chuizheng and Trinh, Loc and Xu, Nan and Enouen, James and Liu, Yan},
  title   = {Interpretability and fairness evaluation of deep learning models on {MIMIC-IV} dataset},
  journal = {Scientific Reports},
  year    = {2022},
  volume  = {12},
  pages   = {7166},
  doi     = {10.1038/s41598-022-11012-2}
}

@article{wang2024readmission,
  author  = {Wang, H. Echo and Weiner, Jonathan P. and Saria, Suchi and Kharrazi, Hadi},
  title   = {Evaluating algorithmic bias in 30-day hospital readmission models: retrospective analysis},
  journal = {Journal of Medical Internet Research},
  year    = {2024},
  volume  = {26},
  pages   = {e47125},
  doi     = {10.2196/47125}
}

@inproceedings{rocheteau2021tpc,
  author    = {Rocheteau, Emma and Li{\`o}, Pietro and Hyland, Stephanie},
  title     = {Temporal Pointwise Convolutional Networks for Length of Stay Prediction in the Intensive Care Unit},
  booktitle = {Proceedings of the Conference on Health, Inference, and Learning (CHIL '21)},
  year      = {2021},
  publisher = {Association for Computing Machinery},
  address   = {New York, NY, USA},
  pages     = {58--68},
  doi       = {10.1145/3450439.3451860}
}

@article{johnson2016mimic,
  author  = {Johnson, Alistair E. W. and others},
  title   = {{MIMIC-III}, a freely accessible critical care database},
  journal = {Scientific Data},
  year    = {2016},
  volume  = {3},
  pages   = {160035},
  doi     = {10.1038/sdata.2016.35}
}

@article{cherian2024fairaudit,
  author  = {Cherian, John J. and Cand{\`e}s, Emmanuel J.},
  title   = {Statistical Inference for Fairness Auditing},
  journal = {Journal of Machine Learning Research},
  year    = {2024},
  volume  = {25},
  number  = {149},
  pages   = {1--49}
}

\end{document}